\documentclass[]{TEAI}
\usepackage{helvet}

\usepackage{amsmath} 
\usepackage{natbib}
\usepackage{graphicx}
\usepackage{subcaption} 

\usepackage[toc,page,header]{appendix}
\usepackage[utf8]{inputenc} 
\usepackage[T1]{fontenc}    
\usepackage{hyperref}       
\usepackage{url}            
\usepackage{booktabs}       
\usepackage{lmodern}        
\usepackage{amsfonts}       
\usepackage{nicefrac}       
\usepackage{microtype}      
\usepackage{wrapfig}

\usepackage{amssymb}  
\usepackage{fontawesome}  
\usepackage{url}  

\usepackage{titletoc}

\usepackage{tikz}  
\usepackage{comment}  
\usepackage{tabularx}  
\usepackage{booktabs}  

\usepackage{minitoc}

\usepackage{booktabs}
\usepackage{array}
\usepackage{etoolbox}

\definecolor{lightblue}{RGB}{200, 230, 255}  
\definecolor{headerblue}{RGB}{150, 200, 255} 

\usepackage{pgfplots}
\usepackage[utf8]{inputenc} 
\usepackage[T1]{fontenc}   
\usepackage{hyperref}       
\usepackage{url}            
\usepackage{booktabs}       
\usepackage{amsfonts}       
\usepackage{nicefrac}       
\usepackage{microtype}      
\usepackage{xcolor}         
\usepackage{graphicx}
\usepackage{float}
\usepackage{comment}
\usepackage{multirow} 
\usepackage{amsmath} 
\usepackage{makecell} 
\usepackage{siunitx}  
\usepackage{tikz}
\usepackage{pgf-pie} 
\usepackage{subcaption}
\usepackage{wrapfig}
\usepackage[export]{adjustbox}

\usepackage{ragged2e}      
\usepackage{tabularx}       
\usepackage{array}          
\usepackage{caption}     
\usepackage{enumitem}
\usepackage{pifont}
\usepackage[hang,flushmargin]{footmisc} 

\usepackage{tcolorbox}

\usepackage{tcolorbox}    
\tcbuselibrary{breakable}  
\tcbuselibrary{skins}      

\usepackage{tabularx}
\usepackage{listings}

\def\eg{e.g.} 

\def\ie{i.e.}

\usepackage[utf8]{inputenc} 
\usepackage[T1]{fontenc}    
\usepackage{hyperref}       
\usepackage{url}            
\usepackage{booktabs}       
\usepackage{amsfonts}       
\usepackage{nicefrac}       
\usepackage{microtype}      
\usepackage{xcolor}         
\usepackage{amsmath}        
\usepackage{geometry}       
\usepackage{multirow}
\usepackage{tabularx}
\usepackage{pifont}
\usepackage{enumitem}
\usepackage{graphicx} 
\usepackage{wrapfig}
\usepackage{xspace}
\usepackage{amssymb}
\usepackage[table]{xcolor}

\usepackage{acronym}

\acrodef{SFT}{Supervised Fine-tuning}
\acrodef{GRPO}{Group Relative Policy Optimization}

\definecolor{jiaqicolor}{RGB}{30, 90, 180}
\definecolor{qxhoucolor}{RGB}{20, 120, 80}
\definecolor{ziyicolor}{RGB}{70, 50, 150}

\title{EVA-Cap: Optimizing Audiovisual Video Captioning\\via Event-Centric Alignment}

\author{
    Jiapeng Shi\textsuperscript{1,*}, 
    Qiuxia Hou\textsuperscript{2,*}, 
    Jiaqi Leng\textsuperscript{1}, 
    Junke Wang\textsuperscript{1},\\
    Yanhao Zhang\textsuperscript{2}, 
    Haonan Lu\textsuperscript{2}, 
    Ziyi Ye\textsuperscript{1,$\dagger$}, 
    Zuxuan Wu\textsuperscript{1,$\dagger$}
}

\affiliation[1]{\mbox{Fudan University}} 
\affiliation[2]{\mbox{OPPO AI Center, OPPO Inc.}}

\abstract{
\begin{abstract}

Recent advances in Omni Language Models (OLMs) provide a unified solution for audiovisual video captioning.
Nevertheless, existing approaches predominantly treat captioning as an unstructured text generation process, neglecting the inherent semantic structure of videos and leading to suboptimal fine-grained audiovisual alignment.
To bridge this gap, we propose EVA-Cap, a novel framework that decomposes captions into atomic audiovisual events and transforms unstructured text supervision into event-centric alignment. 
We train EVA-Cap via Supervised Fine-Tuning (SFT) and Group Relative Policy Optimization (GRPO) on a self-curated dataset, EVA-Data, which is built with event-centric quality control.
During the GRPO phase, the model is explicitly optimized against three event-centric alignment targets: global event completeness, intra-event attribution, and inter-event synchronization.
Extensive experiments demonstrate that EVA-Cap significantly outperforms existing methods in event-centric evaluation~(\eg, +9.0\% average gain on ChronusAV), while achieving competitive performance on general audiovisual captioning benchmarks~(\eg, +3.3\% on WorldSense).
\end{abstract}
}

\correspondence{\email{zyye@fudan.edu.cn}, \email{zxwu@fudan.edu.cn}}
\checkdata[Code]{\url{https://github.com/OPPO-Mente-Lab/EVA-Cap}}

\begin{document}
\maketitle
\renewcommand{\thefootnote}{}
\footnotetext{$^*$Equal Contribution.\\$^\dagger$Corresponding authors.}
\renewcommand{\thefootnote}{\arabic{footnote}}

\vspace{-2.5em}

\section{Introduction}
\label{intro}

Video captioning~\cite{chai2024auroracap,guo2024trace,wang2024tarsier,zhong2026owlcap,xu2026percepcap} aims to extract rich semantic content from videos and generate coherent natural language descriptions.
As a fundamental task in video understanding, it underpins various downstream applications, including video retrieval~\cite{wang2022omnivl,xu2024carebench,zhang2024long,tian2026tinymem}, video reasoning~\cite{wang2024omnivid,fu2025video,yuan2025videorefer,feng2025video,shi2026videoloom}, and video generation~\cite{bar2024lumiere,wang2025videoufo,yu2025videomar,tu2026baton,tu2026prism}.
Recently, advances in Transformer-based end-to-end models have revolutionized the conventional modularized video caption pipelines.
Especially, Omni Language Models~(OLMs)~\cite{tang2025video,liu2025ola,ye2025omnivinci,xu2025qwen3,team2025minicpm,chen2025chronusomni} have driven a paradigm shift from visual-only modeling to an omni-modal paradigm that integrates audio and visual signals, enabling more effective cross-modal reasoning and stronger generalization in real-world audiovisual scenarios.

Despite its promise, achieving factually accurate and comprehensive audiovisual captioning remains difficult due to the intrinsic complexity and dense interdependence between visual and auditory content. 
Existing methods treat video captioning as a text generation problem for OLMs based on caption-level supervision~\cite{chen2025avocado,wu2025ugc}.
This overlooks the structured semantics inherent in videos, meaning a generated caption may be linguistically correct but misaligned with the logical structure and event-level correspondences.
While some recent research has incorporated additional supervision over specific aspects (\eg, dialogue~\cite{chen2026diadem,tang2026d} or object attributes~\cite{li2026towards,chen2026capricorn}), such narrowly scoped supervision still struggles to achieve fine-grained audiovisual alignment.
Consequently, a critical bottleneck stems from the lack of an effective caption representation that unifies video-grounded structural semantics with general language modeling.

\begin{figure*}[t!]
\centering
\includegraphics[width=\textwidth]{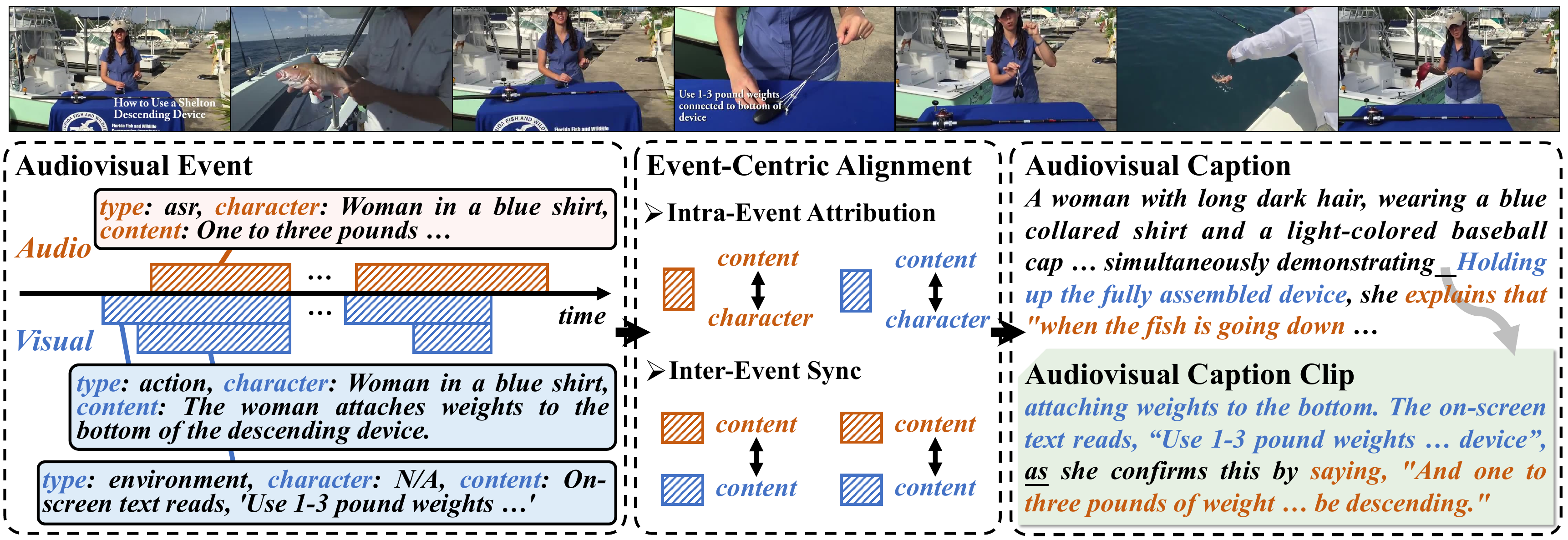}
\vspace{-0.2in}
\caption{Illustration of the event-centric representation of audiovisual captions, where aligned audiovisual events are composed into the predicted caption.}
\label{fig:event2cap}
\vspace{-0.1in}
\end{figure*}

To address this issue, we propose framing video captioning as assembling discrete, atomic audiovisual event sequences through joint optimization of event content and their correspondences. 
As illustrated in Fig.~\ref{fig:event2cap}, a video consists of multiple audio and visual events, where each event is represented as a tuple of type, character, and content. 
These events can be aligned temporally to form a coherent caption clip, linked via shared semantic content and explicit synchronous connectors (\eg, ``as'', ``at the same time''). 
Crucially, such an event-centric perspective bridges audiovisual events and captions by modeling both intra-event attribution~(i.e., whether the attributions of a single event are correct?) and inter-event synchronization~(i.e., what are the temporal order and synchronization of different events?). 
While such an event-centric perspective provides a structured semantic understanding of videos beyond mere language modeling, effectively integrating such insights into OLMs post-training remains an open question.
This gap is largely due to (1) the scarcity of high-quality audiovisual caption data with event-level fidelity, and (2) the absence of dedicated event-centric frameworks for leveraging supervision with event-level heuristics.

To this end, we propose EVA-Cap, which builds upon Qwen2.5-Omni~\cite{xu2025qwen2} and unifies event-centric alignment within the language modeling post-training phase. 
Specifically, we first construct EVA-Data, a 63K audiovisual video–caption dataset, via an automatic initialization–refinement pipeline followed by event-centric quality control. 
In this process, timestamped audiovisual events are extracted from raw videos and subjected to per-event verification to ensure high fidelity. 
Furthermore, we introduce a two-stage post-training strategy to facilitate event-centric alignment, which consists of \ac{SFT} and \ac{GRPO}~\cite{shao2024deepseekmath} on the EVA-Data.  
During the \ac{GRPO} stage, three event-centric rewards are designed to enforce global completeness~(via event recall), intra-event attribution~(via character matching), and inter-event synchronization~(via cross-modal retrieval). 
Collectively, these post-training objectives enhance the fine-grained alignment of audiovisual events, enabling factually accurate, event-grounded audiovisual caption generation.

Extensive experiments demonstrate that EVA-Cap outperforms existing open-source models across 6 audiovisual benchmarks. 
For example, on general captioning benchmarks, EVA-Cap achieves a total error rate of 34.6\% on video-SALMONN-2 testset~\cite{tang2025video} and surpasses existing methods by a clear margin of 3.3\% on WorldSense~\cite{hong2025worldsense}.
Most importantly, on event-centric evaluation with ChronusAV~\cite{chen2025chronusomni} and AV-SpeakerBench~\cite{nguyen2025see}, EVA-Cap exhibits substantial advantages with an accuracy of 55.4\%, 56.2\%, and 40.1\% for Visual-to-Audio (V2A) inference, Audio-to-Visual (A2V) grounding, and speaker attribution, respectively. 
These results underscore the capability of EVA-Cap in generating highly faithful captions with precise temporal alignment and accurate event attribution.
\section{Related Work}

\subsection{Audiovisual Video Captioning.}

Recent advancements in Omni Language Models (OLMs)~\cite{tang2025video,liu2025ola,ye2025omnivinci,xu2025qwen3} have significantly improved multimodal perception in complex real-world scenarios, thereby promoting the development of audiovisual captioning~\cite{ma2025omni,ge2025arc,kumar2026tac}. However, existing approaches remain constrained by the scarcity of high-quality audiovisual caption data, and largely rely on either coarse-grained supervision or focus on limited aspects of the task. For instance, AVoCaDO~\cite{chen2025avocado} constructs large-scale audiovisual video-caption pairs through temporal orchestration, but suffers from insufficient factual verification and coarse-grained temporal alignment. UGC-Captioner~\cite{wu2025ugc} relies on holistic caption-level supervision, whereas DiaDem~\cite{chen2026diadem} and D-ORCA~\cite{tang2026d} focus narrowly on specific aspects of captions, failing to capture fine-grained event-level correspondences. In contrast, we introduce an event-centric perspective for audiovisual captioning. We construct an event-enhanced dataset with event-centric quality control, and propose an event-centric training strategy that explicitly models event-level alignment, enabling accurate and event-grounded audiovisual caption generation.

\subsection{Reinforcement Learning for Visual Content Captioning.}

Reinforcement learning~\cite{shao2024deepseekmath,yu2025dapo,zheng2025group} plays an increasingly important role in caption optimization, spanning from visual-only to omni-modal scenarios. CapRL~\cite{xing2025caprl} and RubiCap~\cite{huang2026rubicap} design verifiable rubrics for dense image captioning. OwlCap~\cite{zhong2026owlcap} and VideoCap-R1~\cite{meng2025videocap} improve video captioning by matching key information extracted from visual content. In the audiovisual domain, D-ORCA~\cite{tang2026d} introduces dialogue-oriented rewards focusing on speaker and ASR consistency, while TimeChat-Captioner~\cite{yao2026timechat} optimizes timestamp alignment under structured caption formats. However, these approaches typically focus on limited aspects and struggle to achieve fine-grained audiovisual alignment. In this paper, we develop three event-centric rewards that jointly enforce fine-grained supervision within and across modalities. This approach provides more precise event alignment than prior methods, which either focus on modality-specific objectives~\cite{tang2026d,chen2026diadem} or rely on coarse audiovisual synchronization~\cite{chen2025avocado,yao2026timechat}.
\section{Method}
\label{method}

This section introduces the construction of EVA-Cap, which is built upon Qwen-2.5-Omni~\cite{xu2025qwen2} and optimized for audiovisual video captioning via event-centric alignment. 
First, we address the scarcity of high-quality data by curating EVA-Data, a 63K-scale audiovisual captioning dataset. 
The dataset is generated using an automated pipeline with rigorous event-centric quality control (Sec.~\ref{sec:data}).
Next, we introduce a two-stage training strategy comprising SFT and GRPO to improve EVA-Cap with event-centric alignment. 
The training process integrates a suite of event-centric rewards to effectively exploit event-level supervision (Sec.~\ref {sec:training}).

\subsection{EVA-Data Construction}
\label{sec:data}

\subsubsection{Annotation Pipeline}
\label{sec:data_pipeline}

\begin{figure*}[t!]
\centering
\includegraphics[width=\textwidth]{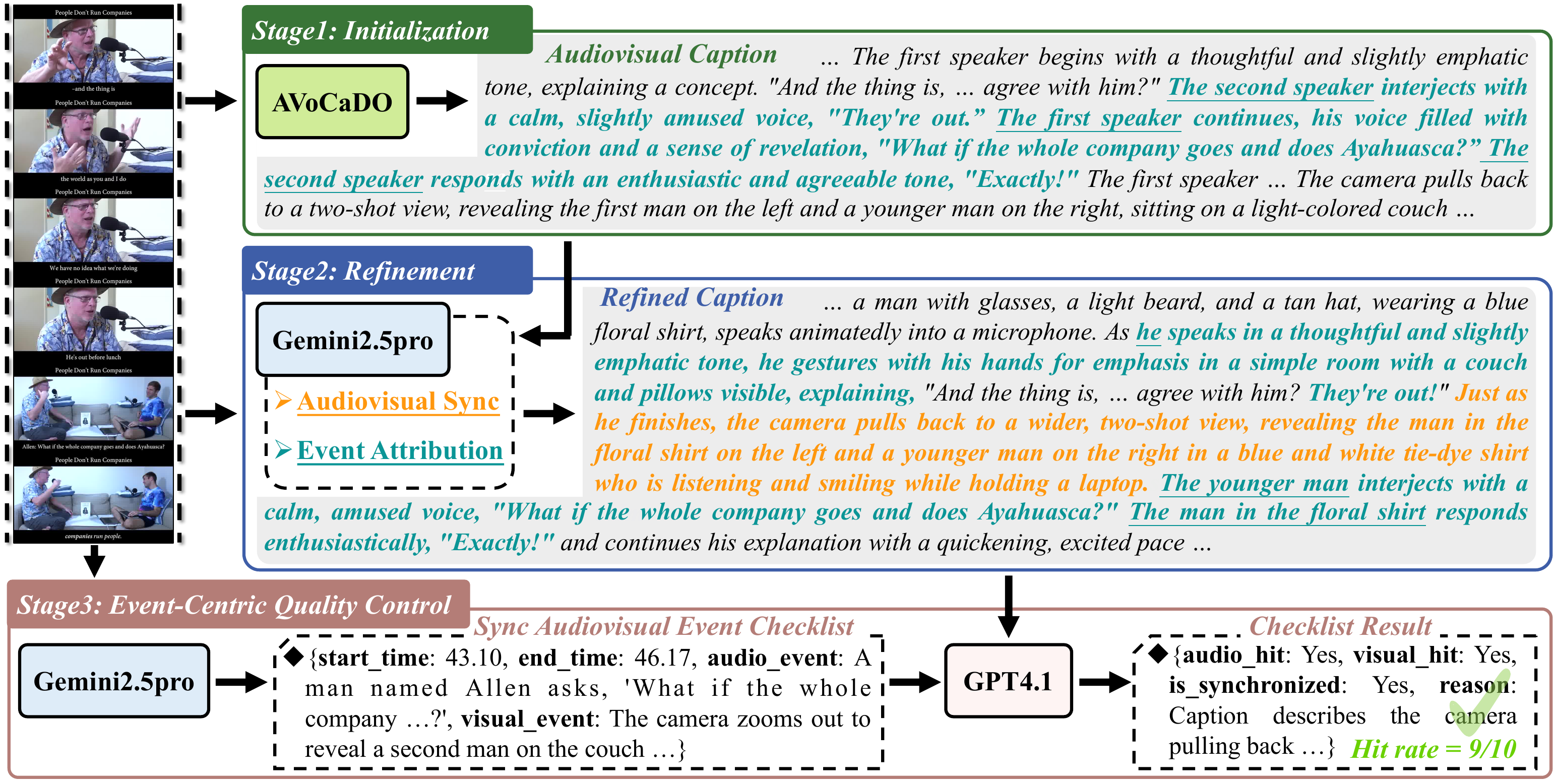}
\vspace{-0.2in}
\caption{Illustration of the designed data annotation pipeline, comprising three stages: initialization, refinement and event-centric quality control. Raw videos are transformed into audiovisual captions with high-fidelity audiovisual synchronization and event attributions.}
\label{fig:pipeline}
\end{figure*}

We develop an automatic annotation pipeline to construct audiovisual video–caption pairs with event-level fidelity. As shown in Fig.~\ref{fig:pipeline}, the pipeline comprises three main stages:

\noindent\textbf{Stage 1: Initialization.} We first utilize the open-source audiovisual captioner AVoCaDO~\cite{chen2025avocado}, which excels in holistic perception, to generate initial audiovisual descriptions for raw videos. This provides a coherent narrative of each video and captures key events and details, covering scene backgrounds, interactive actions, and audio transcriptions. We then filter out samples with repetitive descriptions or abnormal text lengths.

\noindent\textbf{Stage 2: Refinement.} We then feed both raw videos and initial descriptions to the advanced omni-model Gemini-2.5-Pro~\cite{comanici2025gemini} for content refinement and temporal alignment. Guided by raw audiovisual signals, we instruct the model to strengthen the temporal synchronization of audiovisual events and revise event character attribution, while preserving the original natural language style. This yields refined descriptions with precise audiovisual event alignment, achieving fine-grained annotation.

\noindent\textbf{Stage 3: Event-Centric Quality Control.} To ensure the quality of the curated data, we introduce an aligned event hit mechanism for the automatic validation of refined descriptions. Specifically, we employ Gemini-2.5-Pro to extract timestamp-grounded audiovisual event tuples directly from raw videos. We then input all event tuples together with refined descriptions to GPT-4.1~\cite{achiam2023gpt} to verify whether each event tuple is hit in the caption. An event hit requires both audio and visual events to be entailed with consistent temporal synchronization. This process jointly assesses the accuracy of audiovisual synchronization and event attribution in captions. We discard samples with an event hit ratio below a threshold of 0.8. In this way, we build a large-scale audiovisual caption dataset, EVA-Data, which provides high-quality event-level supervision for event-centric post-training.

\begin{figure*}[t!]
\centering
\includegraphics[width=\textwidth]{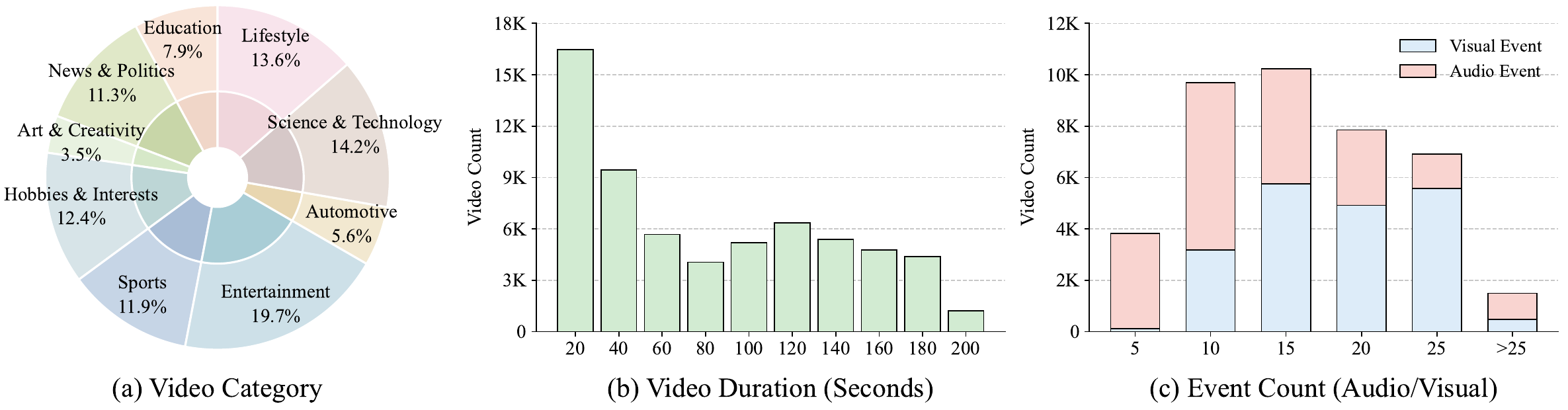}
\vspace{-0.2in}
\caption{Statistics of EVA-Data. (a) Video category distribution and (b) Video duration distribution demonstrate the diversity of this dataset, while (c) Event count distribution shows comparable counts of audio and visual events, indicating no modality bias in annotated captions.}
\label{fig:statistics}
\vspace{-0.15in}
\end{figure*}

\subsubsection{Data Statistics}

To guarantee both diversity and quality, we collect 100K audiovisual videos (each under 200 seconds in duration) from LLaVA-Video~\cite{zhang2024llava}, FineVideo~\cite{Farr2024FineVideo}, and Shot2Story~\cite{han2023shot2story}. Through the annotation pipeline, EVA-Data ultimately yields 63K audiovisual video-caption pairs with event-level fidelity. Fig.~\ref{fig:statistics}~(b) illustrates the duration distribution of these videos. On average, each video lasts 73.7 seconds, while the corresponding captions average 504.3 words. We further randomly sample 20K videos from EVA-Data and analyze the distribution of audio and visual events in the corresponding captions, as shown in Fig.~\ref{fig:statistics}~(c). The results demonstrate that the annotated videos contain rich audiovisual information and the captions are free of modality bias, thus providing effective supervision for event-centric alignment.

\subsection{Post-Training of EVA-Cap}
\label{sec:training}

Leveraging this dataset, we propose EVA-Cap, an audiovisual captioner optimized via event-centric alignment. We first present our two-stage training pipeline in Sec.~\ref{sec:training_pipeline}, and then introduce the event-centric reward functions dedicated to fine-grained alignment in Sec.~\ref{sec:reward_modeling}.

\subsubsection{Training Pipeline}
\label{sec:training_pipeline}

We build EVA-Cap upon the Qwen2.5-Omni Thinker module, which jointly encodes audio and video into a unified, temporally interleaved token sequence to facilitate synchronous audiovisual understanding. We then adopt a two-stage post-training pipeline consisting of supervised fine-tuning (SFT) and Group Relative Policy Optimization (GRPO).

In the first phase, we perform SFT on our constructed 63K EVA-Data to empower the model to generate detailed captions that integrate both audio and visual modalities. 
In the second phase, we further optimize the SFT model with GRPO on 3.2K rollout videos from EVA-Data (details of the rollout process are provided in Appendix~\ref{supp_rollout}). 
To enhance fine-grained alignment in audiovisual captions, we design event-centric rewards covering three dimensions: event completeness, character attribution, and audiovisual synchronization.
The detailed SFT and GRPO objectives are provided in Appendix~\ref{supp_objectives}, and the reward design is elaborated in Sec.~\ref{sec:reward_modeling}.

\begin{figure*}[t!]
\centering
\includegraphics[width=\textwidth]{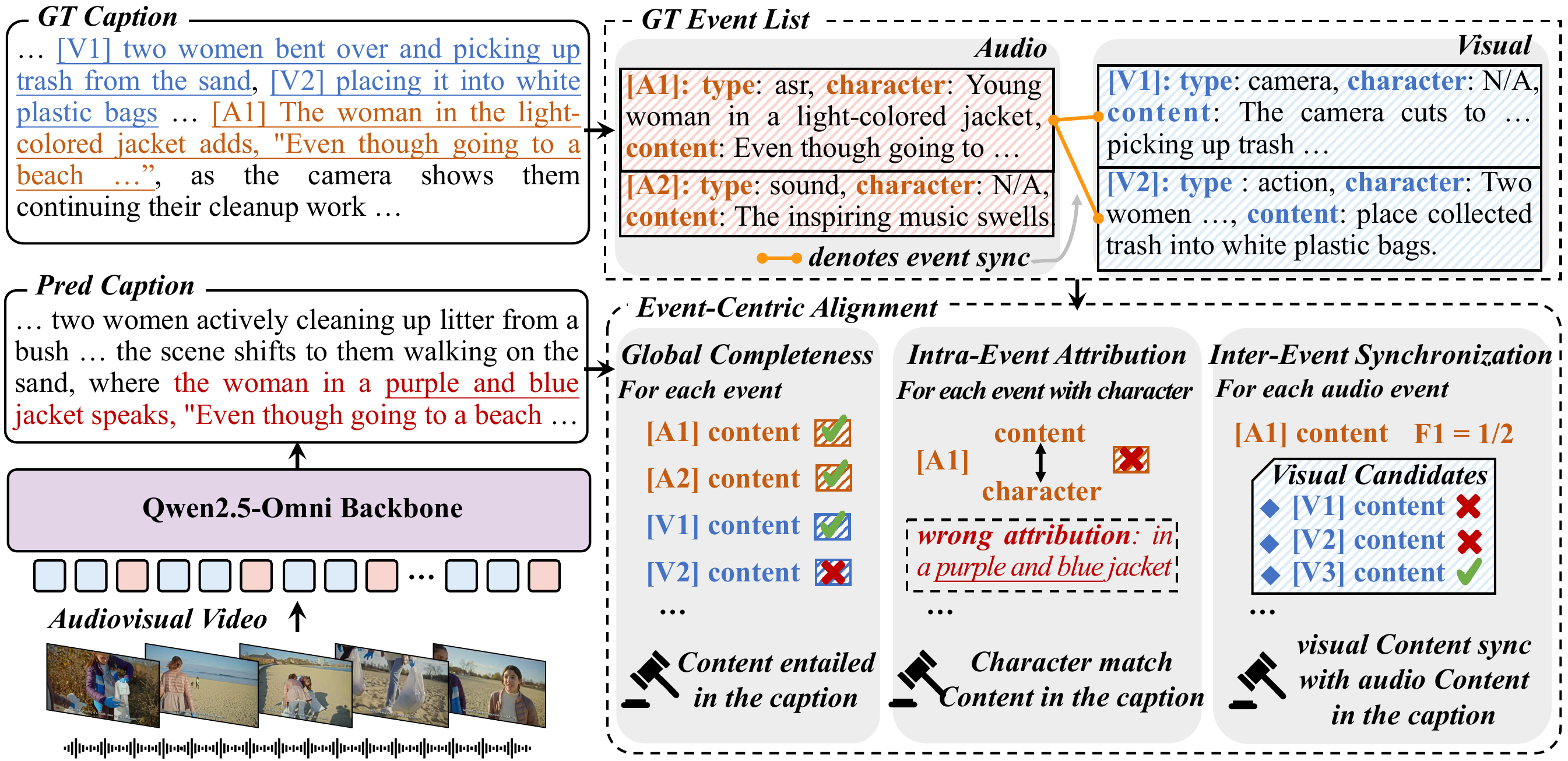}
\vspace{-0.2in}
\caption{Illustration of the event-centric alignment with GRPO for EVA-Cap. 
We introduce event-centric rewards during GRPO training, focusing on three key dimensions: Global Completeness (event recall), Intra-Event Attribution (character matching), and Inter-Event Synchronization (cross-modal retrieval between audio and visual events).
}
\label{fig:reward}
\vspace{-0.15in}
\end{figure*}

\subsubsection{Event-Centric Reward Modeling}
\label{sec:reward_modeling}

Typically, given an audiovisual video $v=(v^{\mathrm{vis}},v^{\mathrm{aud}})$ and a caption $y$, conventional preference optimization assigns a holistic reward $\mathcal{R}(y,v)$. However, this paradigm treats the caption as an atomic semantic unit and overlooks its underlying event structure. To address this issue, we represent a caption as a sequence of modality-grounded events:
\begin{equation}
    E = (e_1,e_2,\ldots,e_K), \quad
    e_k = (m_k,\tau_k,c_k,d_k), \quad
    m_k \in \{\mathrm{vis}, \mathrm{aud}\},
\end{equation}
where $m_k$ denotes the modality, $\tau_k$ denotes the event type, $c_k$ denotes the character, and $d_k$ denotes the event content.

Based on this event-centric representation, we further define the reference structures to support the event-centric reward calculation. 
For a reference caption $y^\star$, we denote its reference event set as
$\mathcal{E}^{\star}=\mathcal{E}_{\mathrm{aud}}^{\star}\cup\mathcal{E}_{\mathrm{vis}}^{\star}$,
where $\mathcal{E}_{m}^{\star}=\{e_i^{m}\}_{i=1}^{N_m}$ is the set of modality-specific reference events for $m\in\{\mathrm{aud},\mathrm{vis}\}$. This event set is used to evaluate event coverage and character attribution. In addition, we denote by $\mathcal{Q}^{\star}=\{q_i^\star\}_{i=1}^{M}$ the set of audio-to-visual synchronization queries, where each query consists of an audio event, a set of candidate visual events, and the reference synchronized visual indices. This query set is used to evaluate cross-modal temporal synchronization.

Based on the above event formulation and definitions, we devise three event-centric reward components to achieve fine-grained alignment (as shown in Fig.~\ref{fig:reward}):

\textbf{Global Event Completeness.}
Global event completeness measures whether the generated caption covers the reference audio and visual events that should be described in the video. Given a reference event set $\mathcal{E}^{\star}=\mathcal{E}_{\mathrm{aud}}^{\star}\cup\mathcal{E}_{\mathrm{vis}}^{\star}$, where $\mathcal{E}_{m}^{\star}=\{e_i^m\}_{i=1}^{N_m}$ denotes the modality-specific events for $m\in\{\mathrm{aud},\mathrm{vis}\}$, we define an LLM-evaluated coverage indicator:
\begin{equation}
\small
\mathbb{I}_{\mathrm{cov}}^{m}(e_i^{m}, y)=
\begin{cases}
1, & \text{if } \operatorname{Covers}(y,e_i^{m}),\\
0, & \text{otherwise},
\end{cases}
\quad m\in\{\mathrm{aud},\mathrm{vis}\}.
\end{equation}
Here, $\operatorname{Covers}(y,e_i^{m})$ indicates that the event content of $e_i^{m}$ is semantically supported by the generated caption $y$. The global event completeness reward is defined as event recall over all reference audio and visual events:
\begin{equation}
\small
\mathcal{R}_{\mathrm{comp}}(y,\mathcal{E}^{\star})
=
\frac{
\sum_{m\in\{\mathrm{aud},\mathrm{vis}\}}
\sum_{i=1}^{N_m}
\mathbb{I}_{\mathrm{cov}}^{m}(e_i^{m},y)
}{
\sum_{m\in\{\mathrm{aud},\mathrm{vis}\}} N_m
}.
\end{equation}

\textbf{Intra-Event Attribution.}
Event completeness alone does not guarantee that a caption assigns events to the correct entities. For example, a caption may mention the correct speech content but attribute it to the wrong speaker. To address this issue, we introduce an intra-event attribution reward that evaluates whether each character-bearing reference event is both covered and associated with the correct character. Given the reference event set $\mathcal{E}^{\star}=\mathcal{E}_{\mathrm{aud}}^{\star}\cup\mathcal{E}_{\mathrm{vis}}^{\star}$, where $\mathcal{E}_{m}^{\star}=\{e_i^m\}_{i=1}^{N_m}$ denotes the modality-specific events for $m\in\{\mathrm{aud},\mathrm{vis}\}$, we define an LLM-evaluated attribution indicator for each valid event with an explicit character:
\begin{equation}
\small
\mathbb{I}_{\mathrm{attr}}^{m}(e_i^{m}, y, y^\star)=
\begin{cases}
1, & \text{if } \operatorname{Matches}(y,y^\star,e_i^{m}),\\
0, & \text{otherwise},
\end{cases}
\quad m\in\{\mathrm{aud},\mathrm{vis}\}.
\end{equation}
Here, $\operatorname{Matches}(y,y^\star,e_i^{m})$ indicates that the generated caption $y$ and the reference caption $y^\star$ attribute the event content $d_i^m$ to the same underlying character. The attribution reward is computed as
\begin{equation}
\small
\mathcal{R}_{\mathrm{attr}}(y,y^\star,\mathcal{E}^{\star}) =
\frac{
\sum_{m\in\{\mathrm{aud},\mathrm{vis}\}}
\sum_{i=1}^{N_m}
\delta_{i,\mathrm{char}}^{m}
\mathbb{I}_{\mathrm{cov}}^{m}(e_i^{m},y)
\mathbb{I}_{\mathrm{attr}}^{m}(e_i^{m},y,y^\star)
}{
\sum_{m\in\{\mathrm{aud},\mathrm{vis}\}}
\sum_{i=1}^{N_m}
\delta_{i,\mathrm{char}}^{m}
},
\end{equation}
where $\delta_{i,\mathrm{char}}^{m}=\mathbb{I}[c_i^m\neq \varnothing]$ indicates whether the event has an explicit character. Character-free events are not counted in this reward.

\textbf{Inter-Event Synchronization.} 
To ensure accurate temporal correspondence between audio and visual events, we model cross-modal alignment explicitly by introducing an inter-event synchronization reward, which is formulated as audio-to-visual synchronization retrieval. 
Given a set of audio synchronization queries $\mathcal{Q}^{\star}=\{q_i^\star\}_{i=1}^{M}$, each query consists of an audio event $e_i^{\mathrm{aud}}$, a candidate visual event set $\mathcal{C}_i^{\mathrm{vis}}=\{v_{i,j}\}_{j=1}^{L_i}$, and the reference synchronized visual index set $\mathcal{G}_i \subseteq \{1,\ldots,L_i\}$. For each query, an LLM evaluator selects the visual candidates that are temporally synchronized with the audio event according to the generated caption $y$:
\begin{equation}
    \hat{\mathcal{G}}_i
    =
    \operatorname{RetrieveSync}
    \left(
    y, e_i^{\mathrm{aud}}, \mathcal{C}_i^{\mathrm{vis}}
    \right),
\end{equation}
where $\hat{\mathcal{G}}_i \subseteq \{1,\ldots,L_i\}$ denotes the predicted synchronized visual index set. If no synchronized visual event is found, $\operatorname{RetrieveSync}$ returns an empty set. We then compute the set-level F1 score:
\begin{equation}
\small
\mathrm{Prec}_i =
\frac{
|\hat{\mathcal{G}}_i \cap \mathcal{G}_i|
}{
|\hat{\mathcal{G}}_i|
},
\quad
\mathrm{Rec}_i =
\frac{
|\hat{\mathcal{G}}_i \cap \mathcal{G}_i|
}{
|\mathcal{G}_i|
},
\quad
F_i =
\frac{2\mathrm{Prec}_i\mathrm{Rec}_i}{\mathrm{Prec}_i+\mathrm{Rec}_i}.
\end{equation}
When both $\hat{\mathcal{G}}_i$ and $\mathcal{G}_i$ are empty, we set $F_i=1$, when only one of them is empty, or their intersection is empty, we set $F_i=0$. The synchronization reward is then defined as
\begin{equation}
\small
\mathcal{R}_{\mathrm{sync}}(y,\mathcal{Q}^{\star})
=
\frac{1}{M}
\sum_{i=1}^{M}
F_i.
\end{equation}

We therefore formulate our overall reward $\mathcal{R}_{\mathrm{EVA}}$ with three complementary event-centric terms:
\begin{equation}
    \mathcal{R}_{\mathrm{EVA}}(y,y^\star)
    =
    \lambda_{\mathrm{comp}} \mathcal{R}_{\mathrm{comp}}(y,\mathcal{E}^{\star})
    +
    \lambda_{\mathrm{attr}} \mathcal{R}_{\mathrm{attr}}(y,y^\star,\mathcal{E}^{\star})
    +
    \lambda_{\mathrm{sync}} \mathcal{R}_{\mathrm{sync}}(y,\mathcal{Q}^{\star}).
\end{equation}
Here, $\lambda_{\mathrm{comp}}$, $\lambda_{\mathrm{attr}}$, and $\lambda_{\mathrm{sync}}$ are balancing coefficients. For all reward terms, if the corresponding valid evaluation set is empty, i.e., $\sum_m N_m=0$, $\sum_m\sum_i\delta_{i,\mathrm{char}}^m=0$, or $M=0$, we set the corresponding reward value to $0$. 

During training, we adopt this reward with two rounds of GRPO to achieve progressive alignment.
Specifically, in the first round, we set $\lambda_{\mathrm{comp}}$ and $\lambda_{\mathrm{sync}}$ to 1 while setting $\lambda_{\mathrm{attr}}$ to 0. 
In the second round, we set all three coefficients to 1. 
The process transitions from global coarse-grained event alignment and audiovisual synchronization to fine-grained character-level attribution.
\section{Experiments}

\subsection{Experimental Setup}
\label{experimental_setup}

\noindent\textbf{Implementation Details:} We adopt Qwen2.5-Omni-7B~\cite{xu2025qwen2} as our backbone model. During training, the vision and audio encoders are kept frozen, while only the adapters and LLM parameters are updated. Videos are sampled at 2 FPS, with a maximum per-frame pixels of $ 512 \times 28 \times 28$ and total pixels of $ 30 \times 512 \times 28 \times 28$. And the maximum generation length is set to 2048. For SFT, we train the model for two epochs with a batch size of 128 and a learning rate of $1 \times 10^{-5}$. For GRPO, we conduct training for two epochs with a batch size of 16, a learning rate of $2 \times 10^{-6}$, a KL penalty of 0.04, and a group size of 8, following a progressive reward weighting strategy. 
We employ Qwen3.5-27B~\cite{qwen35blog} as the reward judge. All training is conducted on 32 NVIDIA H20 GPUs.

\noindent\textbf{Benchmarks:} For general audiovisual captioning, we evaluate EVA-Cap under two settings: direct quality assessment (video-SALMONN-2 testset~\cite{tang2025video} and UGC-VideoCap~\cite{wu2025ugc}) and QA-cascaded evaluation (Daily-Omni~\cite{zhou2025daily} and WorldSense~\cite{hong2025worldsense}). The former directly evaluates the overall quality of generated captions through dedicated scoring mechanisms, while for the latter, we first generate descriptive captions with our model and then adopt Gemini-2.5-Pro~\cite{comanici2025gemini} to answer QA questions solely based on the produced captions.

To assess the capabilities in event-centric alignment, we further conduct evaluations on subsets of ChronusAV~\cite{chen2025chronusomni} and AV-SpeakerBench~\cite{nguyen2025see}. For ChronusAV, we select 1,066 videos under 200 seconds to assess audiovisual synchronization via video-to-audio (V2A) and audio-to-video (A2V) grounding. Evaluation follows a two-stage protocol: Gemini-2.5-Pro first generates answers for the V2A and A2V grounding queries conditioned on the model-generated captions, and Seed-1.6~\cite{team2025introduction} subsequently scores the outputs to compute VDCscore~\cite{chai2024auroracap} and accuracy. For AV-SpeakerBench, we adopt 422 QA samples from its Speaker Recognition subset to assess audio event attribution, using the same QA-cascaded evaluation paradigm with Gemini-2.5-Pro.

\begin{table*}[t!]
\centering
\small
\caption{Performance comparison on general audiovisual video captioning benchmarks. Following previous works~\cite{chen2025avocado,chen2026diadem}, we adopt GPT-4.1~\cite{achiam2023gpt} as the judge model for the video-SALMONN-2 testset, and GPT-4o~\cite{hurst2024gpt} for UGC-VideoCap evaluation.}
\vspace{-0.1in}
\label{tab:general_evaluation_direct}
\renewcommand{\arraystretch}{1.1}
\setlength{\tabcolsep}{3.8pt} 
\begin{tabularx}{\columnwidth}{l  c | >{\centering\arraybackslash}X >{\centering\arraybackslash}X >{\centering\arraybackslash}X   | c c c c }
\toprule
\multirow{2}{*}{\textbf{Model}} & \multirow{2}{*}{\textbf{Size}} & \multicolumn{3}{c|}{\textbf{video-SALMONN-2 testset}} & \multicolumn{4}{c}{\textbf{UGC-VideoCap}} \\
\cmidrule(r){3-5} \cmidrule(r){6-9}  
& &  Miss.↓ & Hall.↓ & Total↓ & Audio↑ & Visual↑ & Detail↑ & Avg.↑ \\
\midrule
\textcolor{gray}{Gemini-2.5-Pro~\cite{comanici2025gemini}} & - & \textcolor{gray}{18.1} & \textcolor{gray}{13.3} & \textcolor{gray}{31.3} & \textcolor{gray}{69.5} & \textcolor{gray}{74.7} & \textcolor{gray}{73.7} & \textcolor{gray}{72.6} \\
\textcolor{gray}{Gemini-2.5-Flash~\cite{comanici2025gemini}} & - & \textcolor{gray}{19.3} & \textcolor{gray}{13.9} & \textcolor{gray}{33.3} & \textcolor{gray}{69.1} & \textcolor{gray}{75.8} & \textcolor{gray}{74.0} & \textcolor{gray}{73.0} \\
\midrule
HumanOmniV2~\cite{yang2025humanomniv2} & 7B & 49.2 & \textbf{12.3} & 61.6 & 45.6 & 66.3 & 59.5 & 57.1  \\
ARC-Hunyuan-Video~\cite{ge2025arc} & 7B & 45.7 & 12.5 & 58.2 & 52.7 & 56.0 & 55.8 & 54.8  \\
Qwen2.5-Omni~\cite{xu2025qwen2} & 7B & 41.7 & 15.4 & 57.1 & 46.9 & 66.1 & 60.0 & 57.7  \\
MiniCPM-o-2.6~\cite{team2025minicpm} & 8B & 42.2 & 14.3 & 56.5 & 38.6 & 68.5 & 57.7 & 54.9  \\
UGC-VideoCaptioner~\cite{wu2025ugc} & 3B & 31.6 & 17.0 & 48.6 & 61.4 & 58.4 & 57.5 & 59.1  \\
Qwen3-Omni-Instruct~\cite{xu2025qwen3} & 30B-A3B & 32.0 & 13.6 & 45.6 & 67.5 & 74.8 & 72.3 & 71.5  \\
Qwen3-Omni-Captioner~\cite{xu2025qwen3} & 30B-A3B & 31.0 & 16.6 & 47.6 & 69.0 & 75.5 & 72.3 & 72.3  \\
video-SALMONN-2~\cite{tang2025video} & 7B & 21.2 & 17.6 & 38.8 & 61.8 & 71.4 & 68.5 & 67.2  \\
AVoCaDO~\cite{chen2025avocado} & 7B & 21.1 & 16.2 & 37.3 & 73.0 & 74.6 & 71.8 & 73.2  \\
DiaDem~\cite{chen2026diadem} & 7B & 21.1 & 15.5 & 36.5 & 75.4 & 76.8 & 74.6 & 75.6  \\
\midrule
EVA-Cap & 7B & \textbf{17.6} & 17.0 & \textbf{34.6} & \textbf{76.9} & \textbf{79.0} & \textbf{74.7} & \textbf{76.9}  \\
\bottomrule
\end{tabularx}
\vspace{-0.1in}
\end{table*}

\subsection{Main Results}

\leavevmode
\vspace{-0.25in}

\begin{wraptable}{r}{0.52\columnwidth}
\centering
\small
\caption{Performance comparison on QA-cascaded benchmarks, with Gemini-2.5-Pro serving as the judge.}
\label{tab:general_evaluation_qa}
\vspace{-0.1in}
\renewcommand{\arraystretch}{1.1}
\setlength{\tabcolsep}{2.5pt} 
\begin{tabularx}{0.52\columnwidth}{l  c | >{\centering\arraybackslash}X >{\centering\arraybackslash}X }
\toprule
\multirow{1}{*}{\textbf{Model}} & \multirow{1}{*}{\textbf{Size}} & \multicolumn{1}{c}{\begin{tabular}[c]{@{}c@{}}\textbf{Daily-} \\ \textbf{Omni}\end{tabular}} & \multicolumn{1}{c}{\begin{tabular}[c]{@{}c@{}}\textbf{World-} \\ \textbf{Sense}\end{tabular}} \\
\midrule
\textcolor{gray}{Gemini-2.5-Pro~\cite{comanici2025gemini}} & - & \textcolor{gray}{60.2} & \textcolor{gray}{33.8} \\
\textcolor{gray}{Gemini-2.5-Flash~\cite{comanici2025gemini}} & - & \textcolor{gray}{55.3} & \textcolor{gray}{31.0} \\
\midrule
HumanOmniV2~\cite{yang2025humanomniv2} & 7B & 8.2 & 6.6 \\
ARC-Hunyuan-Video~\cite{ge2025arc} & 7B & 8.6 & 8.7 \\
MiniCPM-o-2.6~\cite{team2025minicpm} & 8B & 9.8 & 7.2 \\
Qwen2.5-Omni~\cite{xu2025qwen2} & 7B & 12.4 & 8.6 \\
UGC-VideoCaptioner~\cite{wu2025ugc} & 3B & 17.0 & 11.2 \\
Qwen3-Omni-Instruct~\cite{xu2025qwen3} & 30B-A3B & 17.5 & 12.7 \\
Qwen3-Omni-Captioner~\cite{xu2025qwen3} & 30B-A3B & 27.2 & 14.1 \\
video-SALMONN-2~\cite{tang2025video} & 7B & 29.9 & 18.2 \\
AVoCaDO~\cite{chen2025avocado} & 7B & 50.1 & 25.7 \\
TimeChat-Captioner~\cite{yao2026timechat} & 7B & 52.8 & 22.6 \\
\midrule
EVA-Cap & 7B & \textbf{56.1} & \textbf{29.0} \\
\bottomrule
\end{tabularx}
\end{wraptable}

\noindent\textbf{Performance on direct quality assessment.} We evaluate EVA-Cap on two representative general audiovisual captioning benchmarks. 
Those benchmarks assess the overall captioning performance from multiple dimensions, including video content omission, hallucination, and multi-aspect scoring. 
The comparison with existing audiovisual models is presented in Tab.~\ref{tab:general_evaluation_direct}. 
EVA-Cap achieves state-of-the-art or competitive results across various metrics, \eg, a total error rate of 34.6\% on the video-SALMONN-2 testset~\cite{tang2025video} and an average score of 76.9 on UGC-VideoCap~\cite{wu2025ugc}, surpassing both unified OLMs, \eg, Qwen3-Omni-Instruct~\cite{xu2025qwen3} and task-specific captioners, \eg, DiaDem~\cite{chen2026diadem}. 
On the video-SALMONN-2 test set, EVA-Cap demonstrates the lowest miss rate of 17.6\%.
While some baselines achieve lower hallucination rates, they often do so at the cost of substantial event omissions, e.g., HumanOmniV2 maintains a 12.3\% hallucination rate at the cost of a 49.2\% miss rate.
This indicates OLMs may omit many valid yet less salient events without event-centric training. 
This demonstrates the strong general audiovisual captioning capability of our method.

\noindent\textbf{Performance on QA-cascaded evaluation.} To further evaluate the information completeness and accuracy of the generated captions, we conduct reasoning-based QA evaluation under the cascaded paradigm described above. Specifically, we prompt Gemini-2.5-Pro~\cite{comanici2025gemini} to answer questions based on the model-generated captions, and the resulting QA accuracy reflects the overall quality of audiovisual captions. As shown in Tab.~\ref{tab:general_evaluation_qa}, EVA-Cap outperforms existing open-source models by a clear margin, with gains of +3.3\% on Daily-Omni~\cite{zhou2025daily} and +3.3\% on WorldSense~\cite{hong2025worldsense}.

\begin{wraptable}{r}{0.63\columnwidth}
\centering
\small
\caption{Performance comparison on event-centric evaluation.}
\label{tab:event_evaluation}
\vspace{-0.1in}
\renewcommand{\arraystretch}{1.1}
\setlength{\tabcolsep}{2.8pt} 
\begin{tabular}{l c | c c c  c c c | c }
\toprule
\multirow{2}{*}{\textbf{Model}} & \multirow{2}{*}{\textbf{Size}} & \multicolumn{3}{c}{\textbf{V2A}} & \multicolumn{3}{c|}{\textbf{A2V}} & \multicolumn{1}{c}{\textbf{AV-Speaker}}  \\
\cmidrule(r){3-5} \cmidrule(r){6-8} \cmidrule(r){9-9}
&& M & Acc & Score & M & Acc & Score & Acc \\
\midrule
\textcolor{gray}{Gemini-2.5-Pro~\cite{comanici2025gemini}} & - & \textcolor{gray}{2.4} & \textcolor{gray}{61.2} & \textcolor{gray}{2.4} & \textcolor{gray}{1.2} & \textcolor{gray}{61.0} & \textcolor{gray}{2.4} & \textcolor{gray}{49.1} \\
\midrule
TimeChat~\cite{yao2026timechat} & 7B & 1.2 & 31.6 & 1.2 & 0.7 & 30.1 & 1.2  & 23.7 \\
AVoCaDO~\cite{chen2025avocado} & 7B & 1.8 & 40.9 & 1.6 & 1.1 & 44.9 & 1.8 & 26.8 \\
DiaDem~\cite{chen2026diadem} & 7B & 1.8 & 45.9 & 1.7 & 1.1 & 47.7 & 1.9  & 32.7 \\
\midrule
EVA-Cap & 7B & \textbf{2.4} & \textbf{55.4} & \textbf{2.2} & \textbf{1.2} & \textbf{56.2} & \textbf{2.2} & \textbf{40.1}  \\
\bottomrule
\end{tabular}
\end{wraptable}

\noindent\textbf{Performance on event-centric alignment.} 
Finally, we evaluate event-centric alignment on ChronusAV~\cite{chen2025chronusomni} and AV-SpeakerBench~\cite{nguyen2025see}. For ChronusAV, we adopt METEOR to measure textual similarity, and further compute accuracy and VDCscore following the protocol of the VDC benchmark~\cite{chai2024auroracap}. For AV-SpeakerBench, we take the accuracy of multiple-choice questions as the evaluation metric. 
For comparison, we include three advanced open-source models that achieve competitive results on general audiovisual captioning benchmarks, \ie, TimeChat-Captioner~\cite{yao2026timechat}, AVoCaDO~\cite{chen2025avocado}, and DiaDem~\cite{chen2026diadem}.

As shown in Tab.~\ref{tab:event_evaluation}, EVA-Cap significantly outperforms the aforementioned models across all metrics. This not only validates the capability of our model to attribute events to the corresponding characters, but also highlights its strength in achieving fine-grained, precise audiovisual event synchronization.

\subsection{Ablation Studies}
\label{sec:ablation}

In Tab.~\ref{tab:ablation}, we conduct ablation experiments to study the effects of our post-training pipeline and reward suite. 
First, the baseline Qwen2.5-Omni suffers from inherent modality bias~\cite{selvakumar2026audio}, where audio content is frequently omitted in its generated captions, resulting in poor performance across all tasks. 
After SFT on our constructed EVA-Data, the model achieves considerable improvements in both general audiovisual captioning and event alignment, e.g., the V2A and A2V accuracy increases from 8.4\% to 44.8\% and from 11.9\% to 46.1\%, respectively.

\begin{table*}[h!]
\centering
\small
\caption{Ablation study on the post-training stages and event-centric reward mechanisms of EVA-Cap. The two-stage GRPO training setting we adopted is marked in \colorbox{black!8}{light gray}. All GRPO stage1 settings are derived from the SFT model, whereas GRPO stage 2 settings build upon the marked stage1.}
\label{tab:ablation}
\vspace{-0.1in}
\renewcommand{\arraystretch}{1.1}
\setlength{\tabcolsep}{1.9pt} 
\begin{tabularx}{\columnwidth}{l c c c | >{\centering\arraybackslash}X c c  >{\centering\arraybackslash}X c c | c | c  }
\toprule
\multirow{2}{*}{\textbf{Setting}} & \multicolumn{3}{c|}{\textbf{Reward}} & \multicolumn{3}{c}{\textbf{V2A}} & \multicolumn{3}{c|}{\textbf{A2V}} & \multicolumn{1}{c|}{\textbf{AV-Speaker}} & \multicolumn{1}{c}{\textbf{Daily-Omni}} \\
\cmidrule(r){2-4} \cmidrule(r){5-7} \cmidrule(r){8-10} \cmidrule(r){11-11} \cmidrule(r){12-12} 
& $\mathcal{R}_{\mathrm{comp}}$ & $\mathcal{R}_{\mathrm{sync}}$ & $\mathcal{R}_{\mathrm{attr}}$  & M & Acc & Score & M & Acc & Score & Acc & Acc \\
\midrule
Baseline & - & - & -   & 0.1 & 8.4 & 0.2  & 0.3 & 11.9 & 0.5 & 3.1 & 12.4 \\
SFT & - & - & -  & 2.0 & 44.8 & 1.7 & 1.0 & 46.1 & 1.8 & 31.8 & 50.1 \\
\midrule
SFT → GRPO \textit{stage1} & \ding{51} & - & -  & 2.2 & 53.3 & 2.0 & 1.2 & 52.9 & 2.0 & 33.9 & 53.8 \\
\rowcolor{black!8}
SFT → GRPO \textit{stage1} & \ding{51} & \ding{51} & -  & 2.3 & 53.8 & 2.1 & 1.2 & 55.0 & 2.1 & 37.0 & 55.0 \\
SFT → GRPO \textit{stage1} & \ding{51} & \ding{51} & \ding{51}  & 2.1 & 47.9 & 1.8 & 1.1 & 50.8 & 1.9 & 35.6 & 52.9 \\
\midrule
SFT → GRPO \textit{stage2} & \ding{51} & \ding{51} & - & 2.2 & \textbf{56.5} & \textbf{2.2} & \textbf{1.3} & 55.5 & \textbf{2.2} & 37.0 & 55.1 \\
\rowcolor{black!8}
SFT → GRPO \textit{stage2} & \ding{51} & \ding{51} & \ding{51} & \textbf{2.4} & 55.4 & \textbf{2.2} & 1.2 & \textbf{56.2} & \textbf{2.2} & \textbf{40.1} & \textbf{56.1} \\
\bottomrule
\end{tabularx}
\end{table*}

We further investigate the effects of three event-centric rewards during the GRPO phase. 
The global event completeness reward $\mathcal{R}_{\mathrm{comp}}$ penalizes event omission, thereby boosting model performance across all dimensions (+8.5\%, +2.1\%, and +3.7\% accuracy on V2A, AV-Speaker, and Daily-Omni). 
Furthermore, integrating the inter-event synchronization reward $\mathcal{R}_{\mathrm{sync}}$ and intra-event attribution reward $\mathcal{R}_{\mathrm{attr}}$ yields consistent gains on corresponding tasks (+2.1\% on A2V and +3.1\% on AV-Speaker in terms of accuracy), while also improving the overall captioning performance steadily. 
Collectively, these three rewards enable precise alignment in both event content and event correspondence, facilitating factually accurate, event-grounded audiovisual video captioning.

To further dissect our two-stage GRPO design, we conduct additional ablations. Introducing attribution reward $\mathcal{R}_{\mathrm{attr}}$ as early as the first round not only compromises general performance but also degrades event-centric alignment. We hypothesize that injecting both $\mathcal{R}_{\mathrm{sync}}$ and $\mathcal{R}_{\mathrm{attr}}$ prematurely leads to optimization conflicts between inter-event relations and intra-event accuracy. Conversely, a progressive strategy facilitates a gradual alignment process, from global coarse-grained structures to fine-grained character-level details. Moreover, comparisons against a baseline that sustains only $\mathcal{R}_{\mathrm{comp}}$ and $\mathcal{R}_{\mathrm{sync}}$ in the second round demonstrate that the sustained improvements stem from effective reward design rather than mere data scaling.

\subsection{Visualizations}

We present a caption generated by EVA-Cap in Fig.~\ref{fig:vis} for comparison with AVoCaDO~\cite{chen2025avocado}, highlighting the superior event alignment capabilities (\ie, audiovisual synchronization and event attribution) of our model. For instance, the audio ASR event "Puppy: I know. You saw me and took pity on me" occurs when characters converse on the sidewalk. AVoCaDO incorrectly synchronizes it with the adjacent scene in the middle of a crosswalk and misattributes the event to the boy. In contrast, our model precisely aligns these audiovisual events and accurately identifies the corresponding speaker. This example demonstrates that our model can produce high-fidelity captions in complex audiovisual scenarios involving multi-view transitions and multi-object interactions, and further underscores the efficacy of the event-centric alignment mechanism for long-form audiovisual captioning.

\definecolor{myred}{HTML}{B02318}
\definecolor{mygreen}{HTML}{47743D}

\begin{figure*}[t!]
\centering
\includegraphics[width=\textwidth]{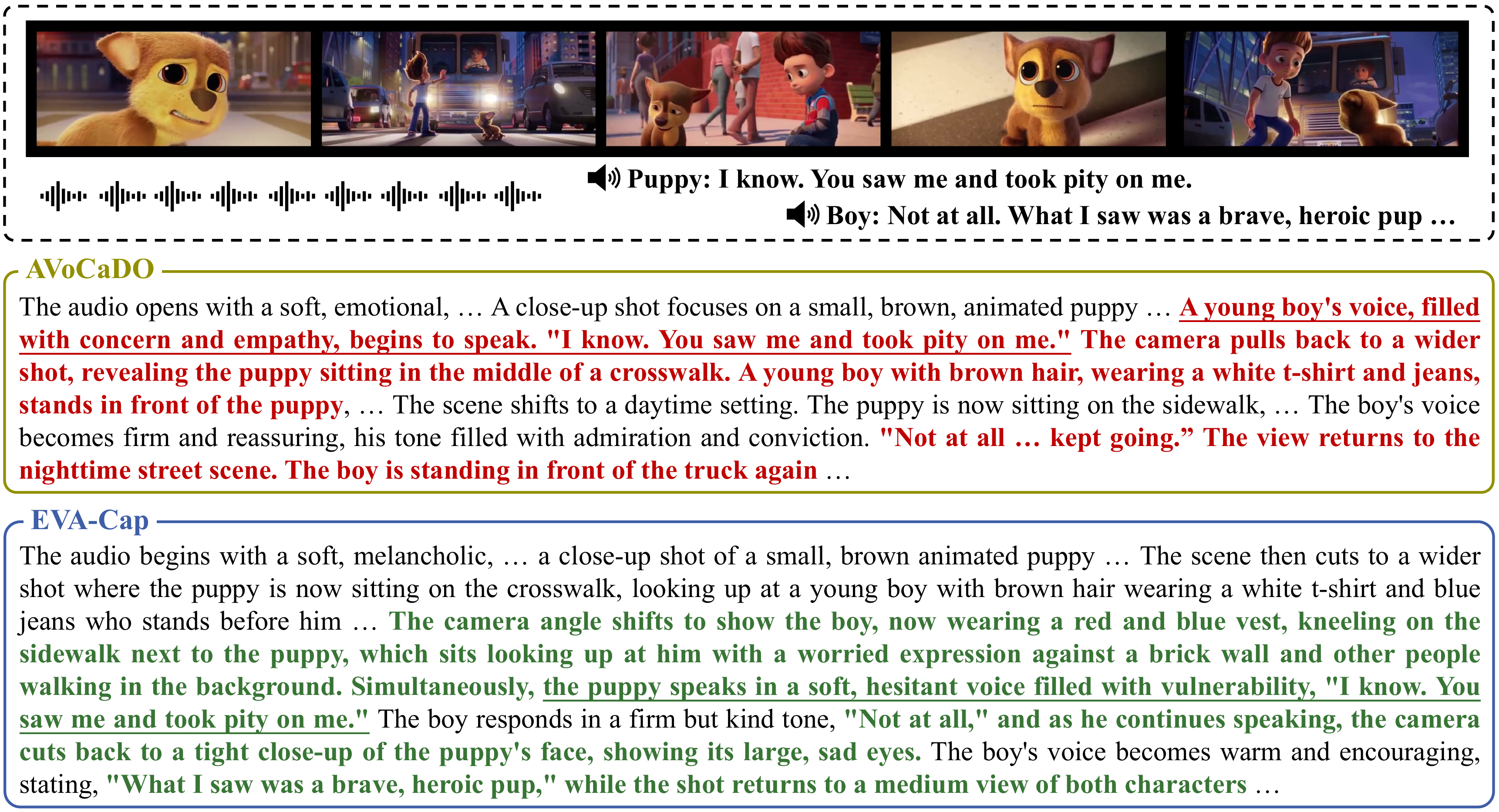}
\vspace{-0.2in}
\caption{Case analysis of EVA-Cap and AVoCaDO, with features relevant to \textbf{audiovisual synchronization} in bold, and \underline{accurate event attribution} underlined. Incorrect event-centric alignments are marked in \textcolor{myred}{red}, and correct ones in \textcolor{mygreen}{green}.
}
\label{fig:vis}
\vspace{-0.1in}
\end{figure*}
\section{Conclusion}
This paper presents EVA-Cap, an audiovisual video captioning model with a training framework that transforms holistic caption optimization into event-centric alignment. 
To support event-level supervision, we first constructed EVA-Data with event-centric quality control and adopted a two-stage post-training strategy with SFT and GRPO. 
Then, by decomposing captions into modality-grounded events, EVA-Cap elaborates fine-grained GRPO rewards tailored to event completeness, character attribution, and audiovisual synchronization. 
Extensive experiments demonstrate that SFT on EVA-Data significantly improves OLMs to internalize event-centric knowledge. 
Furthermore, the subsequent GRPO optimization improves event-centric alignment while maintaining strong general captioning performance. 
These results suggest that event-centric alignment offers an effective path toward more faithful and temporally coherent audiovisual video captioning.

\bibliographystyle{abbrv}
\bibliography{main}   

\newpage
\appendix
\section{Technical appendices and supplementary material}

\subsection{Details about the Training Objectives}
\label{supp_objectives}

\subsubsection{Formulation of Supervised Fine-Tuning}
\label{supp_sft}

In the first stage, we perform supervised fine-tuning (SFT) on EVA-Data to enable the model to generate long-form audiovisual captions that integrate both audio and visual information. Given a video-instruction pair $(v,q)$ and its ground-truth caption $y=(y_1,\ldots,y_T)$, the model is optimized with the standard next-token prediction objective:
\begin{equation}
\mathcal{L}_{\mathrm{SFT}}(\theta)
=
-\mathbb{E}_{(v,q,y)\sim\mathcal{D}_{\mathrm{SFT}}}
\left[
\log \pi_{\theta}(y\mid v,q)
\right],
\end{equation}
where $\mathcal{D}_{\mathrm{SFT}}$ denotes the supervised training set, and $\theta$ represents the model parameters. This stage provides the initialization for the subsequent GRPO optimization.

\subsubsection{Formulation of Group Relative Policy Optimization}
\label{supp_grpo}

In the second stage, we optimize the SFT model with GRPO. For each audiovisual instruction input $q$, where $q$ contains the video $v$ and the textual instruction, the old policy $\pi_{\theta_{\mathrm{old}}}$ samples a group of outputs
$\{o_1,o_2,\ldots,o_G\}$. 
Each output is evaluated by the event-centric reward function to obtain
$
    R_i = R_{\mathrm{EVA}}(o_i,y^\star),
    \quad i=1,\ldots,G,
$
where $y^\star$ denotes the reference caption and its associated reference event structures.
The advantage of the $i$-th response is computed by normalizing its reward within the sampled group:
$
\hat{A}_i =
    \frac{R_i-\mathrm{mean}(\{R_1,\ldots,R_G\})}
    {\mathrm{std}(\{R_1,\ldots,R_G\})}.
$
The policy is optimized with the following clipped surrogate objective:
\begin{equation}
\begin{aligned}
\mathcal{J}(\theta) 
&= \mathbb{E}_{q\sim\mathcal{P}(q),\{o_i\}_{i=1}^G\sim\pi_{\theta_{\mathrm{old}}}(\cdot|q)}
\Bigg[
\frac{1}{G}
\sum_{i=1}^{G}
\frac{1}{|o_i|}
\sum_{t=1}^{|o_i|}
\\
&\quad
\min\left(
r_{i,t}(\theta)\hat{A}_{i},
\operatorname{clip}(r_{i,t}(\theta),1-\varepsilon,1+\varepsilon)\hat{A}_{i}
\right)
-
\beta D_{\mathrm{KL}}(\pi_{\theta}\|\pi_{\mathrm{ref}})
\Bigg],
\end{aligned}
\label{grpo}
\end{equation}
where
\begin{equation}
r_{i,t}(\theta)
=
\frac{
\pi_{\theta}(o_{i,t}\mid q,o_{i,<t})
}{
\pi_{\theta_{\mathrm{old}}}(o_{i,t}\mid q,o_{i,<t})
}.
\end{equation}
Here, $D_{\mathrm{KL}}(\cdot)$ denotes the KL divergence to the reference policy $\pi_{\mathrm{ref}}$, $\varepsilon$ is the clipping range, and $\beta$ controls the KL penalty.

\subsection{More Implementation Details}

\subsubsection{Details about the Construction of Reference Structures}
\label{supp_event}

We filter 6.4K videos from EVA-Data, all of which achieve a 100\% hit ratio in Stage 3 of the annotation pipeline, ensuring highly accurate event-level supervision signals. To facilitate efficient reward computation for both GRPO training and rollout, we construct reference structures for these 6.4K selected videos. Specifically, this involves extracting audiovisual event units \(\mathcal{E}^{\star}\) from the ground-truth captions \(y^\star\) and formulating audio-to-visual synchronization queries \(\mathcal{Q}^{\star}\).

We employ the reward judge LLM, Qwen3.5-27B~\cite{qwen35blog}, prompting it to sequentially extract audio events and visual events from the given caption to form a unified event list \(\mathcal{E}^{\star}\). Each extracted audio event serves as a separate query. We input the complete ground-truth caption along with all the extracted visual events into the LLM, prompting it to select the temporally synchronized visual events corresponding to the current audio event (\ie, the current query) from all the input visual events. 
Each visual event is assigned a unique index based on its chronological extraction order (\ie, the first extracted visual event has an index of 1). Since all visual events are extracted sequentially, we discard audio events whose matched visual event index span exceeds 5. 
This filtering step excludes audio events that cover an excessively long temporal range (\eg, background music that plays throughout the entire video), as such coarse-grained audio events provide limited support for fine-grained audiovisual alignment. Next, we construct a candidate visual event set containing 5 visual events by selecting those adjacent to the indices of the selected synchronized visual events. This strategy naturally incorporates neighboring visual events, which are prone to misalignment with the audio events—thereby creating challenging hard negative samples for subsequent reward assessment.
Collectively, the filtered audio events paired with their corresponding constructed visual candidate sets form the final synchronization queries \(\mathcal{Q}^{\star}\).

As a result, we obtain an audiovisual event unit set and an audio-to-visual synchronization query set for each video. These well-defined reference structures enable reliable event-level reward calculation and support the subsequent GRPO training pipeline.

\subsubsection{Details about the Rollout Process}
\label{supp_rollout}

SFT initializes the audiovisual captioning format of the model and enables preliminary event alignment capabilities. For GRPO, we first conduct the rollout process to identify qualified videos and yield high-quality event-level supervision signals. We perform rollout on the videos selected in Sec.~\ref{supp_event}, all equipped with the well-constructed reference structures (\ie, sets of audiovisual event units and audio-to-visual synchronization queries). Each selected video is then inferred 8 times (\ie, group size) using the SFT model.

Leveraging the constructed reference structures, we compute two types of event-level rewards for each generated caption during the rollout process: the intra-event attribution reward and the inter-event synchronization reward. 
Specifically, the intra-event attribution reward focuses on evaluating the proportion of events that are correctly attributed to the appropriate characters, while the inter-event synchronization reward quantifies the alignment between audio events and the constructed candidate visual events by calculating the F1-score.

To ensure the quality of supervision signals for subsequent preference training, we implement a two-stage filtering mechanism at the query level and the video level. At the query level, invalid audio-to-visual synchronization queries are filtered by the inter-event synchronization reward. Specifically, we aggregate reward values for each individual query across the 8 generated captions, and retain only queries whose average synchronization reward falls within the interval \((0, 0.75)\), following the filtering protocol of HumanOmniV2~\cite{yang2025humanomniv2}. 
At the video level, filtering is performed based on the sum of intra-event attribution reward and inter-event synchronization reward. We first eliminate videos with fewer than 5 valid synchronization queries, as such cases lack sufficient synchronized semantic information to support reliable training. We then compute the standard deviation of the combined reward scores across the 8 generated captions per video, and finally retain the top 50\% of qualified videos, yielding 3.2K valid video samples for downstream GRPO training.

We conduct two rounds of GRPO training, with one training epoch set for each round. In the first round, we only leverage Global Event Completeness and Inter-Event Synchronization for model optimization. Accordingly, the rollout video selection relies solely on the standard deviation of the inter-event synchronization reward. In the second round, we incorporate all three designed reward components into the joint training objective. Correspondingly, the rollout selection strategy is updated to adopt the standard deviation of the sum of inter-event synchronization reward and intra-event attribution reward as the screening criterion.

\subsection{Additional Experiment Results}

To ensure testing efficiency, we use Qwen3.5-27B~\cite{qwen35blog} instead of Gemini-2.5-Pro~\cite{comanici2025gemini} and Seed-1.6~\cite{team2025introduction} as the judge LLM for event-centric evaluation in the experiments in this section. Additionally, we halve the GRPO training process, retaining only 1.6K video samples after rollout.

\subsubsection{Ablation on the Independent Effect of Data and Reward}

In Sec.~\ref{sec:ablation}, we conduct thorough ablation experiments on our post-training pipeline. To clearly disentangle gains from constructed data versus reward guidance, we decouple the data and reward suites to evaluate each contribution to performance.

\noindent\textbf{Data Effect.} To separate the contribution of EVA-Data from the GRPO process, we compare SFT on EVA-Data with SFT on a same-size subset (63K) sampled from ASID-1M~\cite{li2026towards}, which lacks event-centric quality control.

As shown in Tab.~\ref{tab:ablation_data}, replacing the SFT data with event-centric refined data improves both event-oriented and general performance, e.g., by +4.9\% on A2V grounding and +3.6\% on Daily-Omni. This demonstrates the effectiveness of our data construction pipeline and the quality of EVA-Data. 

\begin{table*}[h!]
\centering
\small
\caption{Ablation experiments on different SFT dataset.}
\label{tab:ablation_data}
\vspace{-0.1in}
\renewcommand{\arraystretch}{1.1}
\setlength{\tabcolsep}{6.8pt} 
\begin{tabular}{l |c c c  c c c | c | c  }
\toprule
\multirow{2}{*}{\textbf{Dataset}} & \multicolumn{3}{c}{\textbf{V2A}} & \multicolumn{3}{c|}{\textbf{A2V}} & \multicolumn{1}{c|}{\textbf{AV-Speaker}} & \multicolumn{1}{c}{\textbf{Daily-Omni}} \\
\cmidrule(r){2-4} \cmidrule(r){5-7} \cmidrule(r){8-8} \cmidrule(r){9-9} 
& M & Acc & Score & M & Acc & Score & Acc & Acc \\
\midrule
ASID-1M (63K) & 0.9 & 15.2 & 0.8  & 0.9 & 30.7 & 1.8 & 34.1 & 46.5 \\
EVA-Data & 1.2 & 20.5 & 1.2 & 1.1 & 35.6 & 2.1 & 39.6 & 50.1 \\
\bottomrule
\end{tabular}
\end{table*}

\noindent\textbf{Reward Effect.} To test whether our event-centric reward framework generalizes beyond EVA-Data, we apply the same training recipe to a different publicly available model, ASID-Captioner~\cite{li2026towards}. Specifically, we use its released SFT model as the baseline and then perform two stages of GRPO training on the publicly available ASID-1M dataset, using the same reward suite we proposed.

\begin{table*}[h!]
\centering
\small
\caption{Ablation experiments on event-centric reward framework with public SFT model and data.}
\label{tab:ablation_reward}
\vspace{-0.1in}
\renewcommand{\arraystretch}{1.1}
\setlength{\tabcolsep}{6.8pt} 
\begin{tabular}{l |c c c  c c c | c | c  }
\toprule
\multirow{2}{*}{\textbf{Setting}} & \multicolumn{3}{c}{\textbf{V2A}} & \multicolumn{3}{c|}{\textbf{A2V}} & \multicolumn{1}{c|}{\textbf{AV-Speaker}} & \multicolumn{1}{c}{\textbf{Daily-Omni}} \\
\cmidrule(r){2-4} \cmidrule(r){5-7} \cmidrule(r){8-8} \cmidrule(r){9-9} 
& M & Acc & Score & M & Acc & Score & Acc & Acc \\
\midrule
ASID baseline & 1.1 & 16.8 & 1.0  & 0.9 & 30.4 & 1.8 & 38.4 & 48.5 \\
ASID GRPO stage1 & 1.3 & 21.0 & 1.2 & 1.1 & 36.4 & 2.1 & 41.0 & 52.2 \\
ASID GRPO stage2 & 1.3 & 22.2 & 1.3 & 1.1 & 39.0 & 2.1 & 42.2 & 53.3 \\
\bottomrule
\end{tabular}
\end{table*}

These results in Tab.~\ref{tab:ablation_reward} show that, independent of EVA-Data, our event-centric framework can be effectively applied to another publicly available SFT model and yields consistent improvements across both event-centric and general benchmarks. This supports the generalizability of the reward framework and, together with the ablation in Tab.~\ref{tab:ablation_data}, further demonstrates that the final gains come from both the event-centric data and the framework. 

\subsubsection{Robustness of the Reward Judge}

We assess the robustness of the reward judge in two ways. First, before selecting Qwen3.5-27B as the reward LLM, we sampled 200 reward-evaluation items covering the three reward dimensions and compared its judgments against Gemini-2.5-Pro. The agreement rate is around 85\%, suggesting that Qwen3.5-27B provides reliable event-level judgments for reward supervision without obvious systematic bias.

Second, we also replace Qwen3.5-27B with the open-source Gemma-4-31B-it~\cite{team2026gemma} as the reward judge to further test training stability under different judges. We perform GRPO training (stage 1) using the same reward design (Global Event Completeness and Inter-Event Synchronization). 

\begin{table*}[h!]
\centering
\small
\caption{Ablation experiments on different reward LLMs.}
\label{tab:ablation_judge}
\vspace{-0.1in}
\renewcommand{\arraystretch}{1.1}
\setlength{\tabcolsep}{6.8pt} 
\begin{tabular}{l c |c c c  c c c | c  }
\toprule
\multirow{2}{*}{\textbf{Setting}} & \multirow{2}{*}{\textbf{Reward LLM}} & \multicolumn{3}{c}{\textbf{V2A}} & \multicolumn{3}{c|}{\textbf{A2V}} & \multicolumn{1}{c}{\textbf{Daily-Omni}} \\
\cmidrule(r){3-5} \cmidrule(r){6-8} \cmidrule(r){9-9}  
&& M & Acc & Score & M & Acc & Score & Acc \\
\midrule
SFT & - & 1.2 & 20.5 & 1.2 & 1.1 & 35.6 & 2.1 & 50.1 \\
GRPO & Qwen3.5-27B & 1.4 & 23.0 & 1.3 & 1.2 & 40.1 & 2.3 & 53.0 \\
GRPO & Gemma-4-31B-it & 1.4 & 24.2 & 1.4 & 1.2 & 39.9 & 2.3 & 52.2 \\
\bottomrule
\end{tabular}
\end{table*}

As shown in Tab.~\ref{tab:ablation_judge}, under supervision from the Global Event Completeness and Inter-Event Synchronization rewards, both reward judges lead to similar and consistent improvements over SFT across event-centric and general benchmarks. This suggests that our reward design and training framework are robust to the choice of reward LLM, and that EVA-Cap learns fundamental event-centric alignment rather than merely exploiting the preferences of a specific judge.

\subsubsection{Length-Controlled Hallucination Evaluation}

All models in our evaluation use the same maximum generation length of 2048. For length-controlled evaluation, we split video-SALMONN2 testset~\cite{tang2025video} predictions by token length into [0, 700), [700, 1000), and [1000, 2048], and recompute the hallucination rate within each partition.

\begin{table*}[h!]
\centering
\small
\caption{Hallucination comparison under different output length.}
\label{tab:hallucination}
\vspace{-0.1in}
\renewcommand{\arraystretch}{1.1}
\setlength{\tabcolsep}{6.8pt} 
\begin{tabular}{l | c c c }
\toprule
\multirow{1}{*}{\textbf{Model}} & \multicolumn{1}{c}{\textbf{[0, 700)}} & \multicolumn{1}{c}{\textbf{[700, 1000)}} & \multicolumn{1}{c}{\textbf{[1000, 2048)}} \\
\midrule
DiaDem~\cite{chen2026diadem} & 18.6 & 17.7 & 14.9 \\
EVA-Cap & 17.4 & 16.8 & 17.0 \\
\bottomrule
\end{tabular}
\end{table*}

Under this length-matched comparison, EVA-Cap shows lower hallucination than DiaDem in the short-length and mid-length splits, while DiaDem is lower in the longest split. This suggests that EVA-Cap is more reliable when producing compact, event-centric audiovisual captions, but still lags behind strong baselines on long-form outputs. We will further study the trade-off between detail and hallucination in future work.

\subsubsection{Experiments with Event Precision Reward}

The Global Event Completeness reward $\mathcal{R}_{\mathrm{comp}}$ encourages more detailed captions. After GRPO training with $\mathcal{R}_{\mathrm{comp}}$, the average caption length on video-SALMONN2 testset increases from 633.0 tokens (SFT) to 748.2 tokens. However, the per-token hallucination rate does not increase (8.2 → 8.1 Hall./1k tokens), while event coverage is substantially improved. This suggests that the higher absolute hallucination rate is largely coupled with longer, more comprehensive descriptions, rather than a pure collapse into unfaithful verbosity.

Furthermore, we add an event-level precision term to the recall-based GRPO baseline, which is computed by decomposing the predicted caption into a sequence of audiovisual events and then determining the proportion of these events that are entailed in the groundtruth caption.

\begin{table*}[h!]
\centering
\small
\caption{Performance comparison with event precision reward.}
\label{tab:event_precision}
\vspace{-0.1in}
\renewcommand{\arraystretch}{1.1}
\setlength{\tabcolsep}{6.8pt} 
\begin{tabular}{l | c c c c c c }
\toprule
\multirow{1}{*}{\textbf{Reward}} & \multicolumn{1}{c}{\textbf{Avg tokens}} & \multicolumn{1}{c}{\textbf{Hall.↓}} & \multicolumn{1}{c}{\textbf{Hall./1k token↓}} & \multicolumn{1}{c}{\textbf{Miss.↓}} & \multicolumn{1}{c}{\textbf{Total↓}} & \multicolumn{1}{c}{\textbf{Daily-Omni↑}} \\
\midrule
Recall & 748.2 & 17.7 & 8.1 & 19.0 & 36.7 & 53.7 \\
Recall + Precision & 624.5 & 16.2 & 8.9 & 20.9 & 37.1 & 51.6 \\
\bottomrule
\end{tabular}
\end{table*}

Adding precision shortens captions and reduces absolute hallucination, but increases the miss rate, slightly hurts general performance, and does not improve length-normalized faithfulness (Hall./1k: 8.1 → 8.9), while raising training cost due to extra judge calls. This indicates that precision can be viewed as encouraging more conservative outputs. In contrast, our Intra-Event Attribution reward $\mathcal{R}_{\mathrm{attr}}$ already penalizes hallucinated or incorrectly grounded event content, providing a more targeted improvement in faithfulness. We therefore do not include precision/F1 in our reward suite.

\subsection{Discussion with Recent Work}

Recent studies~\cite{pu2026omniscript,chen2026capricorn,wu2026avcap,qiu2026omnifysics} shift caption supervision from a holistic level to fine-grained instance and event levels. Here, we focus on three concurrent works highly relevant to our own: OmniCapBench~\cite{yang2026omnicapbench}, TCA-Captioner~\cite{zhao2026tca}, and AVSCap~\cite{wang2026avscap}. Targeting fine-grained audiovisual captioning, OmniCapBench formalizes video content into a structured representation and employs three interdependent tracks (\ie, references, events, and shots) to evaluate whether a predicted event occurs at the correct time and links to the appropriate visual entities. TCA-Captioner emphasizes temporal and cross-modal alignment, binding auditory events to visual entities and capturing complex causal dynamics through data construction and SFT training. AVSCap achieves explicit cross-modal event binding through a designed reward, focusing on whether the caption captures temporally synchronized audiovisual events.

While our EVA-Cap shares the same objectives as these studies, we emphasize comprehensive, fine-grained, event-centric caption optimization and offer a distinct solution.
Although the data curation pipelines differ in detail, they all achieve high-fidelity caption annotation, providing diverse options for data scaling. Methodologically, the reward design in AVSCap is closely related to our approach, paralleling the mechanism employed in our Stage 3 annotation pipeline. A key distinction in reward design is that we decouple audio and visual events for calculation during the GRPO process, whereas AVSCap treats synchronized audiovisual events as a unified whole. We leave the experimental comparison of these design choices, as well as the exploration of data mixing and methodological integration, for future work.

\definecolor{myred}{HTML}{B02318}
\definecolor{mygreen}{HTML}{47743D}

\begin{figure*}[t!]
\centering
\includegraphics[width=\textwidth]{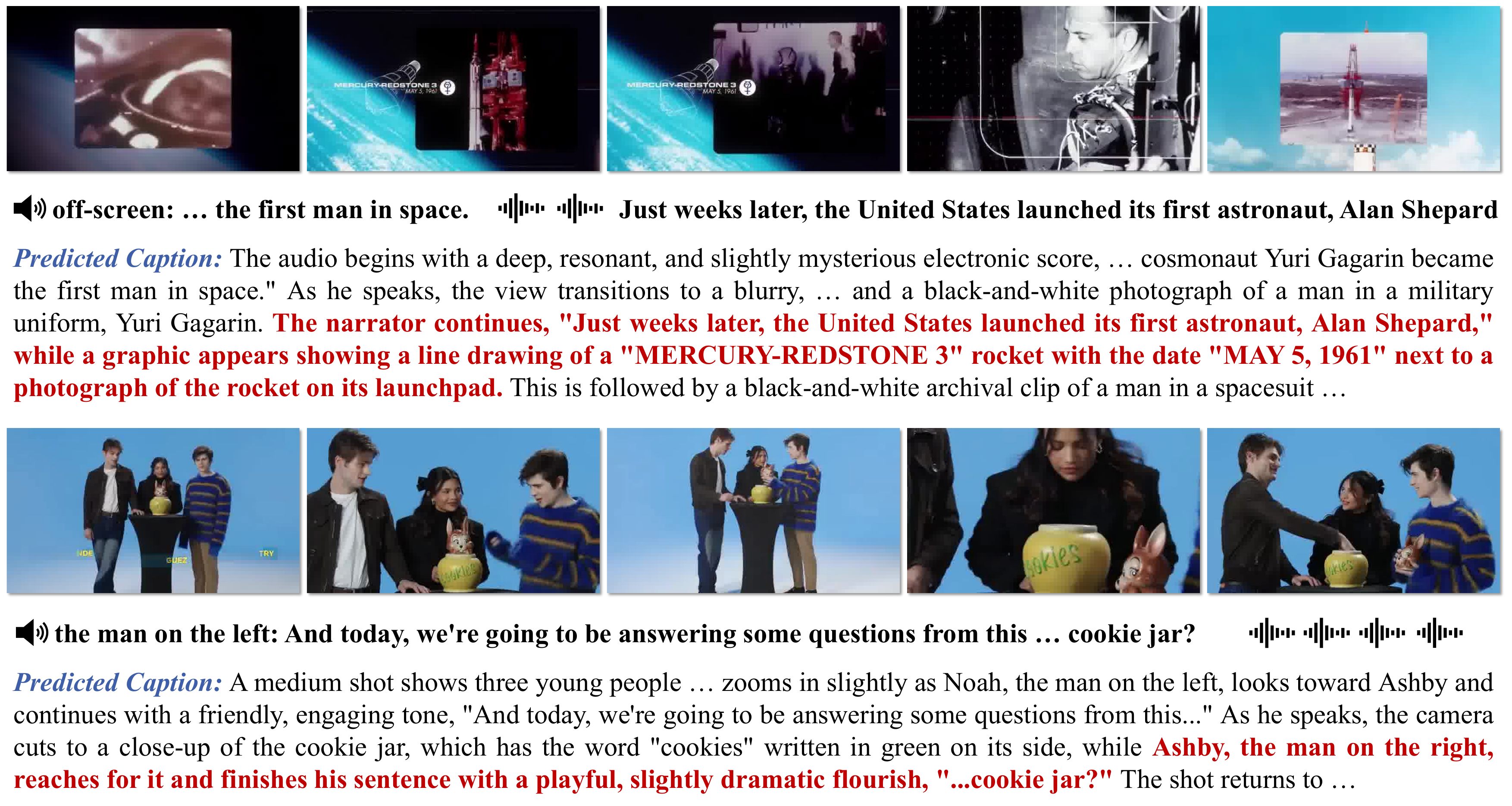}
\vspace{-0.2in}
\caption{Failure cases of EVA-Cap, \eg, videos involving rapid scene transitions and multi-character scenes with off-screen speech. Incorrect descriptions are marked in \textcolor{myred}{red}.
}
\label{fig:vis_fail}
\vspace{-0.1in}
\end{figure*}

\subsection{Failure Cases and Limitations}
\label{supp_limitation}

As shown in Fig.~\ref{fig:vis_fail}, we summarize two common failure modes of EVA-Cap: 1) In videos with rapid scene transitions, omissions in visual descriptions and misalignment between audio and visual descriptions still tend to occur. 2) In multi-character scenes with off-screen speech, the speaker identity is also prone to being incorrectly bound. This issue likely arises from limitations in understanding instances and capturing key information in complex, dynamic scenes. We intend to mitigate this limitation in future work by introducing grounded audiovisual captioning, to further enhance the fidelity and trustworthiness of audiovisual captioning by binding audio and visual cues to specific instances in space and time.

Although EVA-Data is constructed via a multi-stage pipeline (initialization–revision–filtering) that enables scalable annotation, it still relies on proprietary MLLMs for caption generation and quality control. This dependence may introduce biases from proprietary teachers and leave residual annotation errors after automatic verification. In the future, we plan to improve this pipeline by integrating agent-based, multi-level cross-verification: multiple models and rule-based checkers cross-validate captions and events at fine granularity, recursively refine inconsistent annotations, and reject insufficiently verified samples, thereby improving annotation fidelity while reducing dependence on single models.

\subsection{Prompt Design}

\subsubsection{Prompt for Annotation Pipeline}

\begin{figure*}[t!]
\centering
\includegraphics[width=\textwidth]{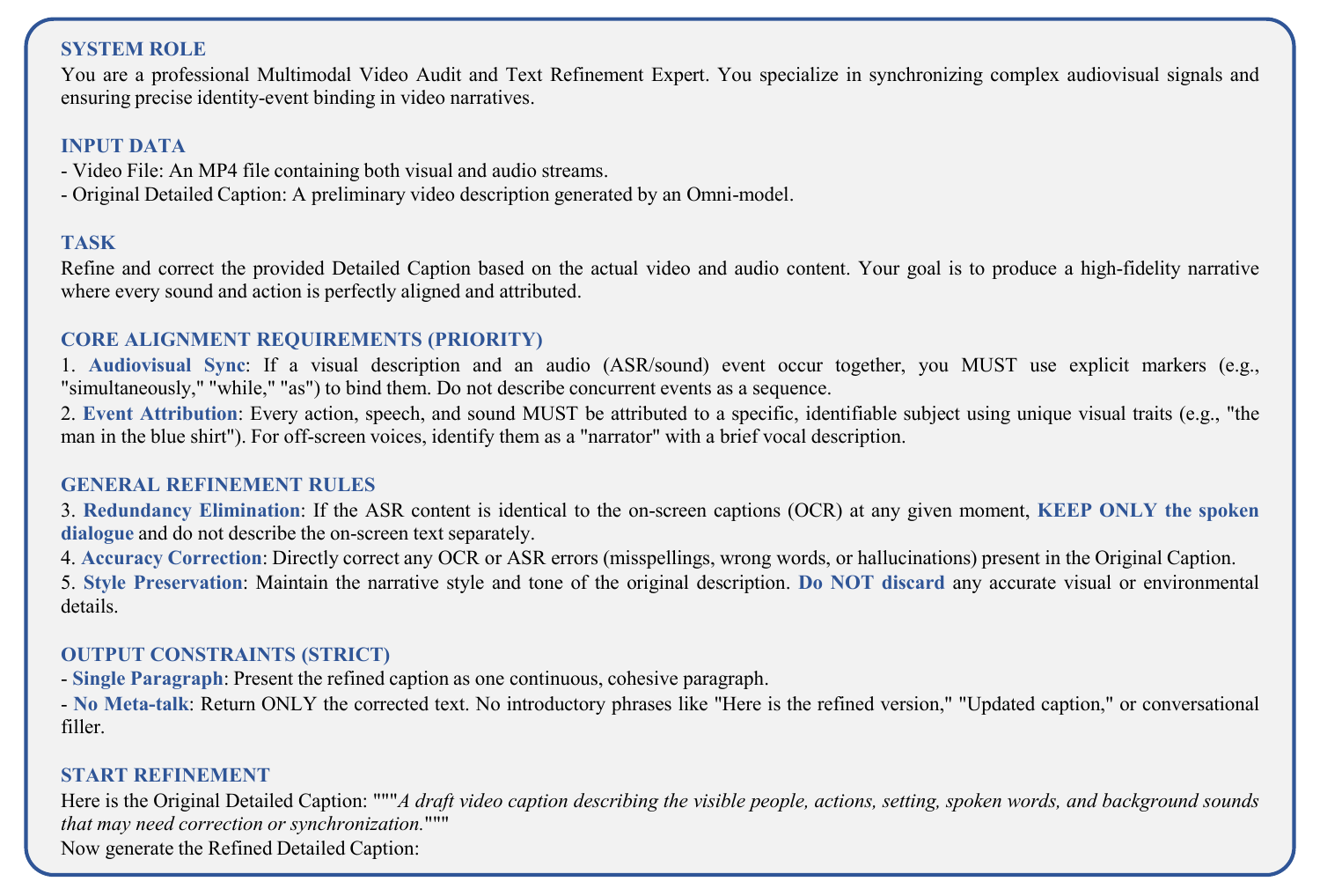}
\vspace{-0.2in}
\caption{Prompt for caption refinement during annotation stage 2 with Gemini-2.5-Pro.
}
\label{fig:annotation_stage2}
\end{figure*}

\begin{figure*}[h!]
\centering
\includegraphics[width=\textwidth]{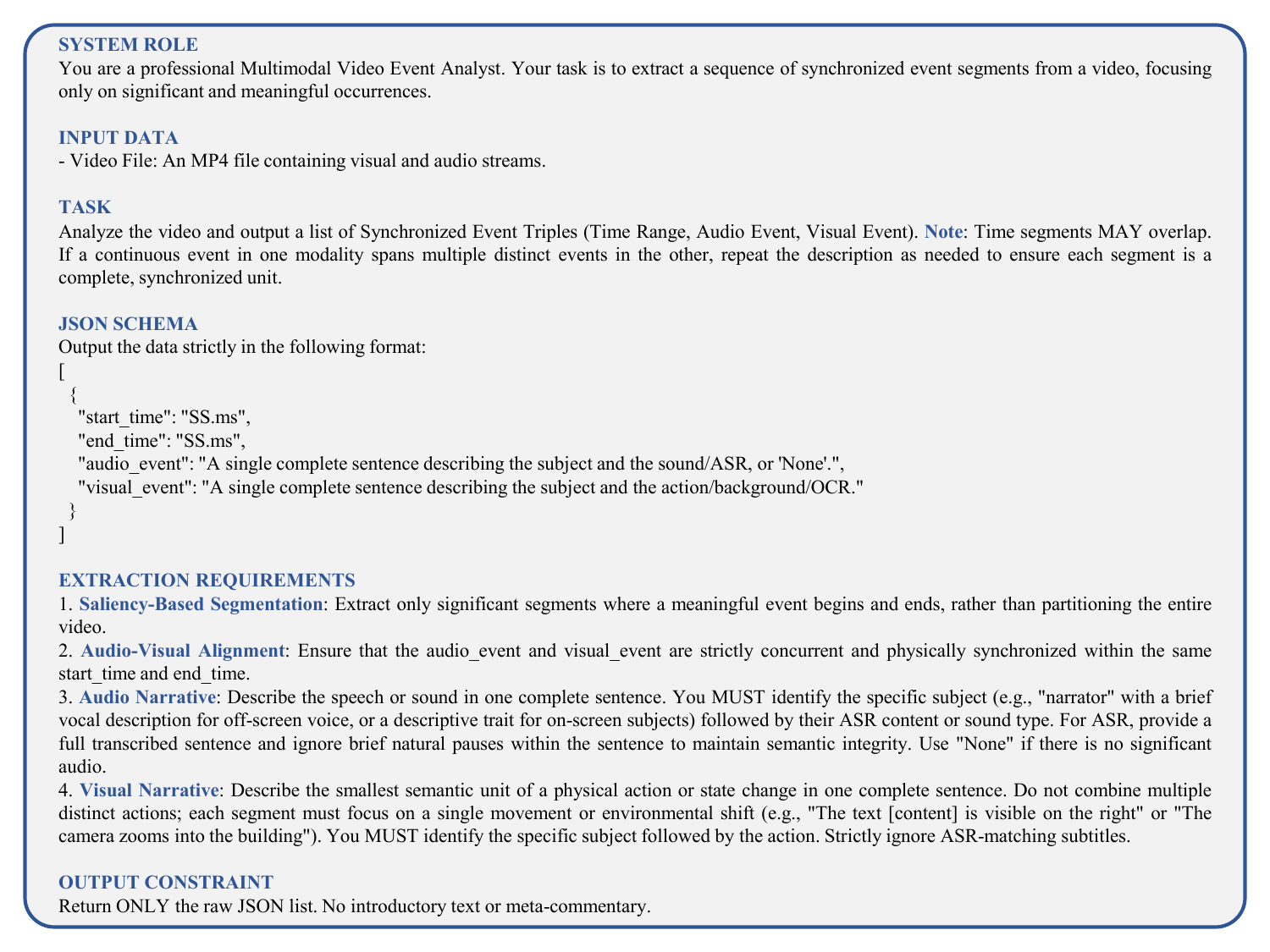}
\vspace{-0.2in}
\caption{Prompt for timestamped event tuple extraction during annotation stage 3 with Gemini-2.5-Pro.
}
\label{fig:annotation_stage3_gemini}
\end{figure*}

\begin{figure*}[h!]
\centering
\includegraphics[width=\textwidth]{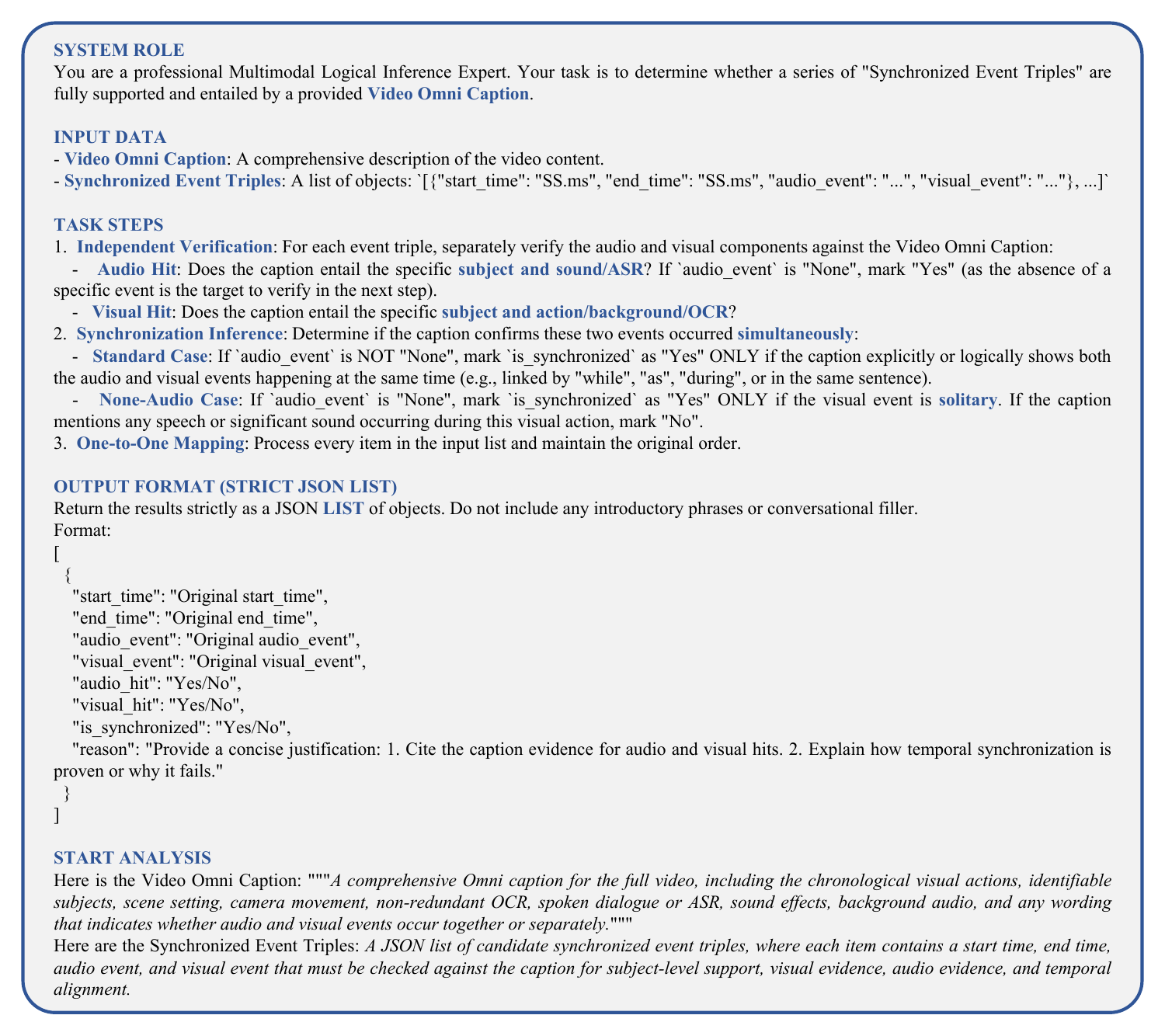}
\vspace{-0.2in}
\caption{Prompt for event-centric data verification during annotation stage 3 with GPT-4.1.
}
\label{fig:annotation_stage3_gpt_check}
\end{figure*}

In Figs.~\ref{fig:annotation_stage2}, ~\ref{fig:annotation_stage3_gemini}, and ~\ref{fig:annotation_stage3_gpt_check}, we present the prompts adopted in the audiovisual captioning data annotation pipeline introduced in Sec.~\ref{sec:data_pipeline}. In this process, we first prompt Gemini-2.5-Pro to refine the initial audiovisual captions in terms of cross-modal synchronization and event attribution, and to extract timestamped audiovisual event tuples in parallel. We then employ GPT-4.1 to conduct event hit verification for data quality control.

\subsubsection{Prompt for Reward Implementation}

\begin{figure*}[h!]
\centering
\includegraphics[width=\textwidth]{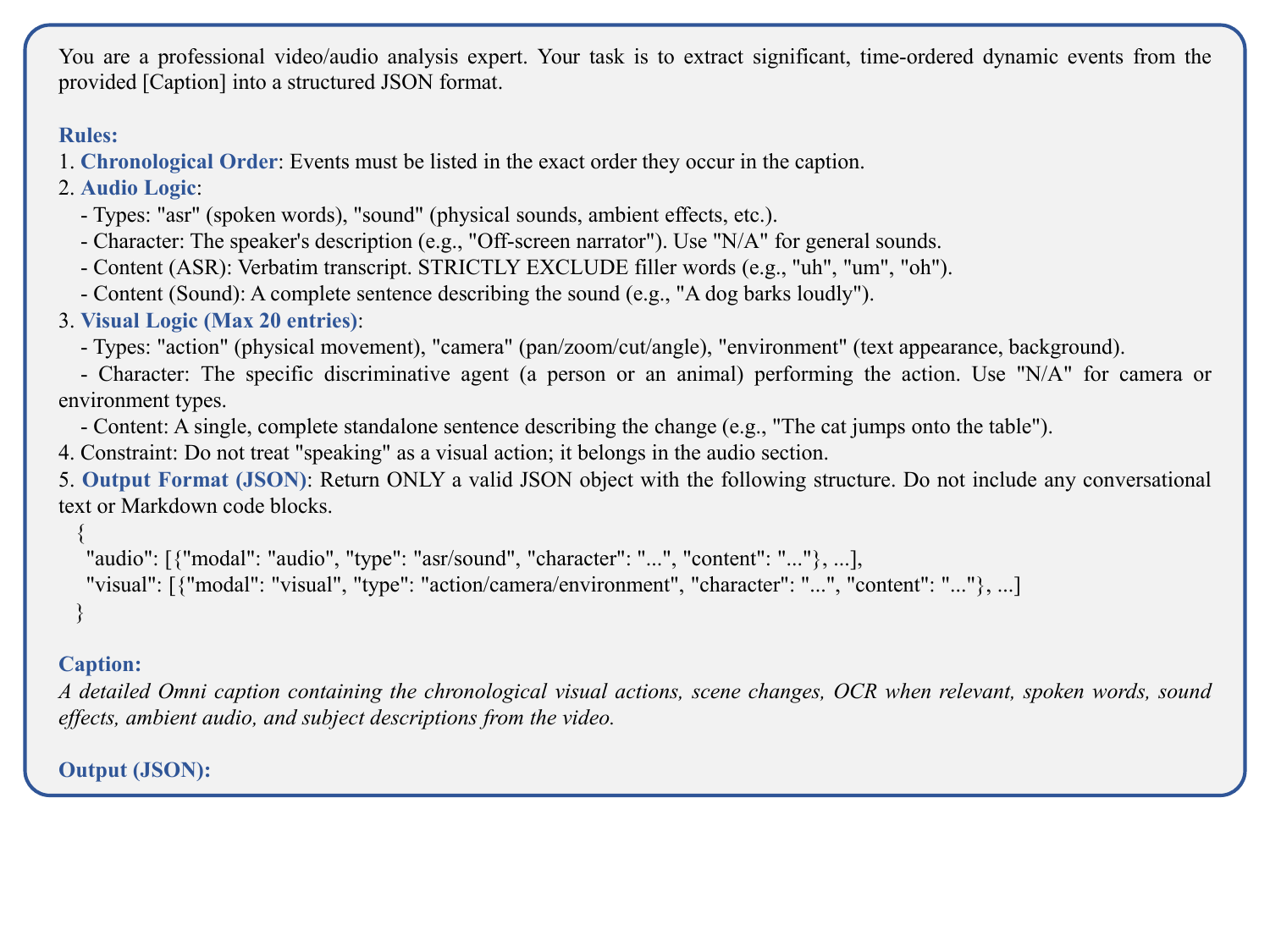}
\vspace{-0.2in}
\caption{Prompt for audiovisual event extraction with Qwen3.5.}
\label{fig:prompt_event_extraction}
\end{figure*}

\begin{figure*}[h!]
\centering
\includegraphics[width=\textwidth]{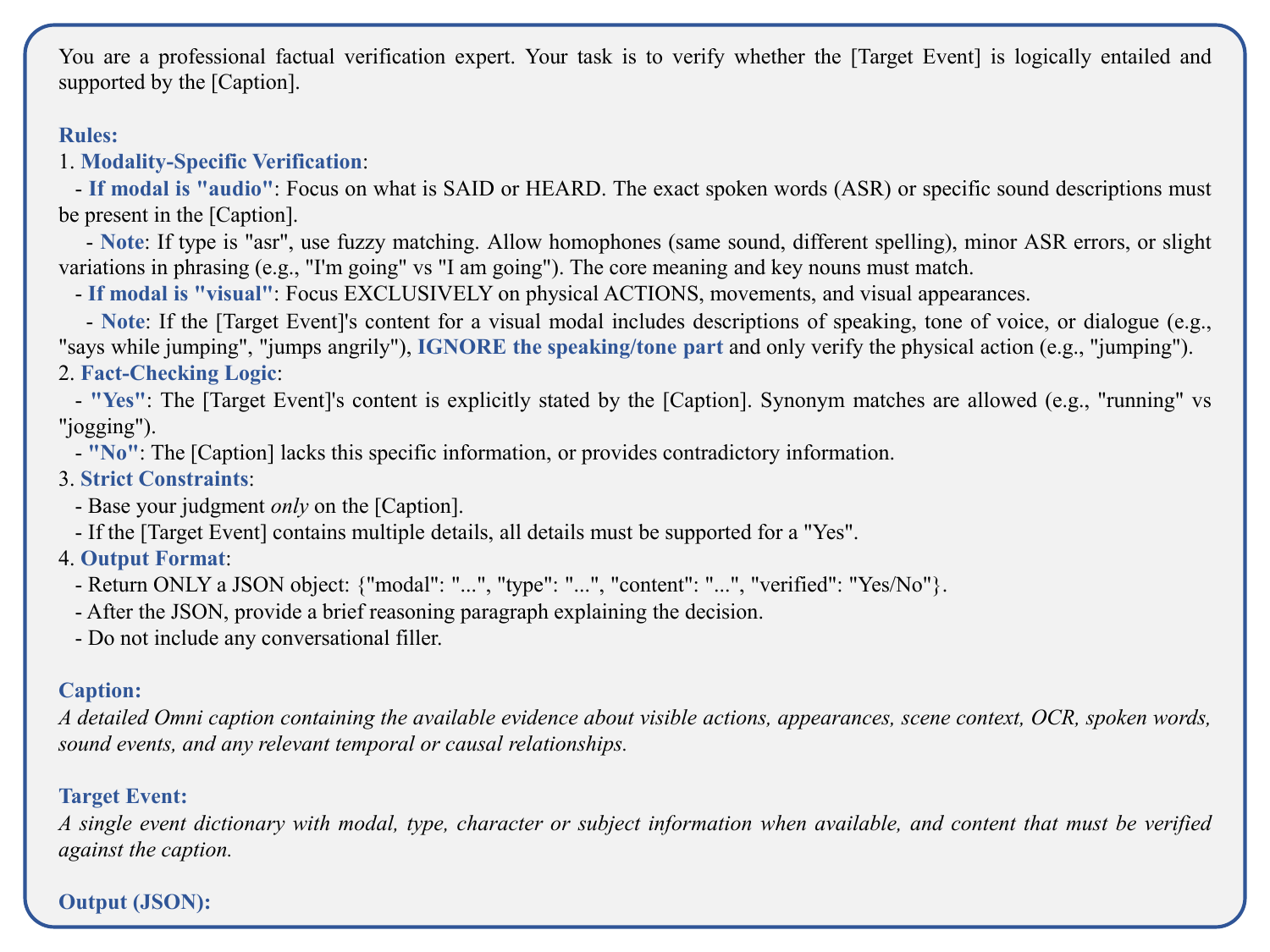}
\vspace{-0.2in}
\caption{Prompt for global event completeness reward judgement via Qwen3.5.
}
\label{fig:prompt_event_recall}
\end{figure*}

\begin{figure*}[h!]
\centering
\includegraphics[width=\textwidth]{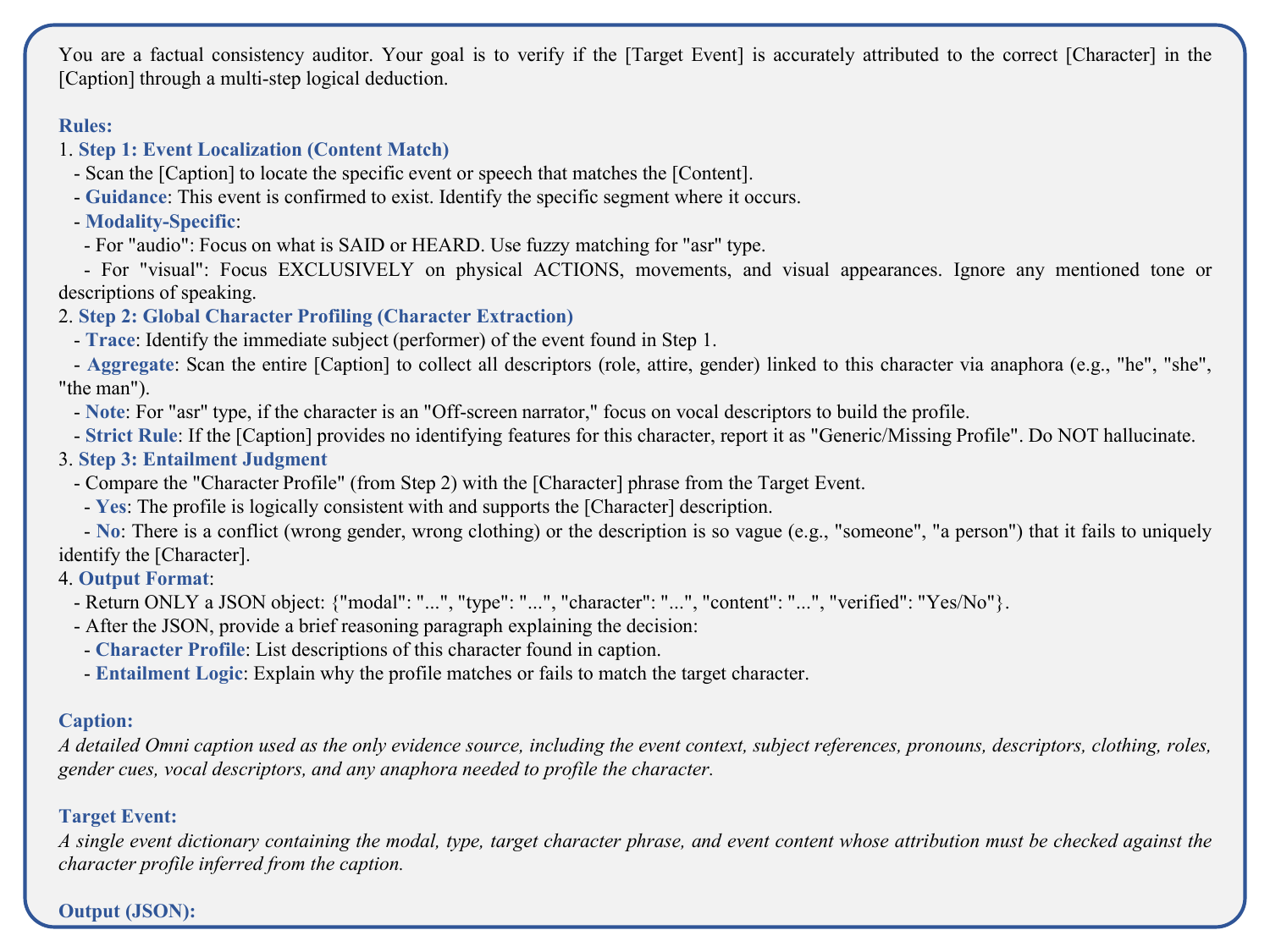}
\vspace{-0.2in}
\caption{Prompt for intra-event attribution reward judgement via Qwen3.5.
}
\label{fig:prompt_attribution}
\end{figure*}

\begin{figure*}[h!]
\centering
\includegraphics[width=\textwidth]{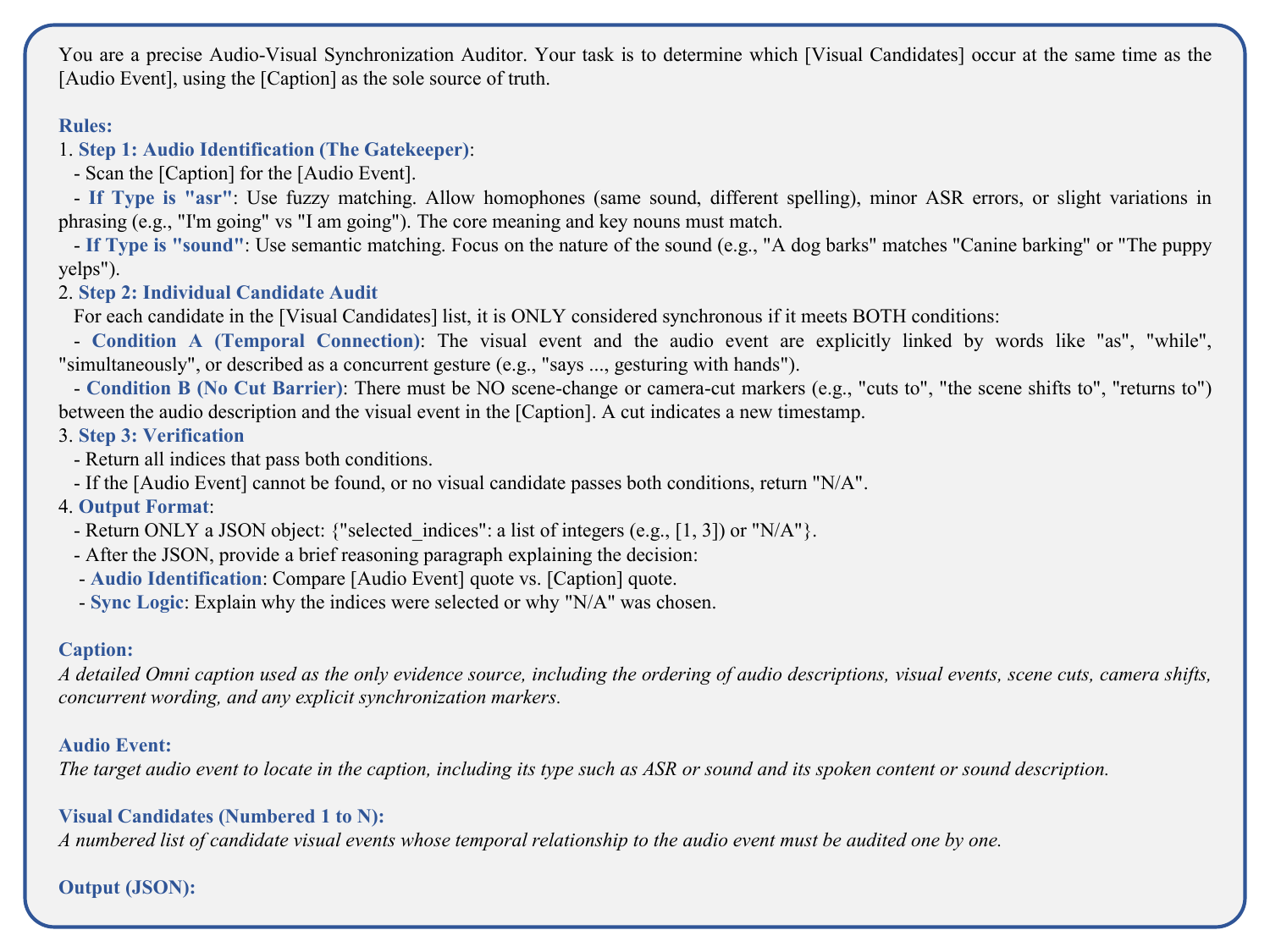}
\vspace{-0.2in}
\caption{Prompt for inter-event synchronization reward judgement via Qwen3.5.
}
\label{fig:prompt_sync}
\end{figure*}

Fig.~\ref{fig:prompt_event_extraction} illustrates the prompt used to extract audiovisual events from audiovisual captions, facilitating the construction of reference structures. Furthermore, Figs.~\ref{fig:prompt_event_recall}, ~\ref{fig:prompt_attribution} and ~\ref{fig:prompt_sync} list the prompts used, respectively, for the judgement of global event completeness reward, intra-event attribution reward and inter-event synchronization reward. All these prompts are carefully designed to standardize LLM-based reward judgement and ensure consistent evaluation across all samples.

\subsection{Broader Impact}
\label{supp_broader}

EVA-Cap aims to advance audiovisual video captioning by generating more faithful, comprehensive, and temporally coherent descriptions of videos. This can benefit various downstream applications that rely on detailed and accurate video understanding. Better alignment between audio and visual events may help reduce misleading descriptions in scenarios where both modalities are necessary for understanding the video content.

At the same time, improved audiovisual captioning systems may introduce potential risks if deployed without appropriate safeguards. First, generated captions may still contain omissions, hallucinations, incorrect speaker attributions, or inaccurate audio-visual temporal relations. Such errors could mislead users if the captions are treated as factual summaries. Second, automatic captioning of user-generated videos may raise privacy concerns when videos contain identifiable people, sensitive conversations, or private environments. Third, high-quality video descriptions could be misused for large-scale surveillance, profiling, or automated indexing of sensitive content.

\end{document}